\documentclass{article}

\usepackage[unreviewed]{neurips_2021}

\usepackage[utf8]{inputenc} %
\usepackage[T1]{fontenc}    %
\usepackage{xcolor}         %
\usepackage[colorlinks,linkcolor={blue!50!black},citecolor={blue!50!black},urlcolor={blue!50!black}]{hyperref}       %
\usepackage{url}            %
\usepackage{booktabs}       %
\usepackage{amsfonts}       %
\usepackage{nicefrac}       %
\usepackage{microtype}      %

\usepackage{amsmath,amssymb}
\usepackage{graphicx}
\usepackage{tabularx}
\usepackage{enumerate}
\usepackage[frozencache]{minted}
\usepackage{stmaryrd}
\usepackage{subfigure}

\renewcommand{\vec}[1]{{\mathbf{#1}}}
\newcommand{\vx}{{\vec x}}

\newcommand{\vX}{{\vec X}}

\newcommand{\vXX}{{\boldsymbol {\mathcal X}}}
\newcommand{\XX}{{\mathfrak X}}
\newcommand{\ZZ}{{\mathfrak Z}}
\newcommand{\EE}{{\mathbb E}}

\newcommand{\eq}{\!=\!}

\newcolumntype{Y}{>{\centering\arraybackslash}X}

\title{Understanding Entropy Coding With Asymmetric Numeral Systems (ANS): a Statistician's Perspective}

\author{%
  Robert Bamler \\
  University of Tübingen \\
  Department of Computer Science \\
  Maria-von-Linden-Straße 6 \\
  72076 Tübingen, Germany \\
  \texttt{robert.bamler@uni-tuebingen.de} \\
}

\begin{document}

\maketitle

\begin{abstract}
Entropy coding is the backbone data compression.
Novel machine-learning based compression methods often use a new entropy coder called Asymmetric Numeral Systems (ANS)~\citep{duda2015use}, which provides very close to optimal bitrates and simplifies~\citep{townsend2019practical} advanced compression techniques such as bits-back coding.
However, researchers with a background in machine learning often struggle to understand how ANS works, which prevents them from exploiting its full versatility.
This paper is meant as an educational resource to make ANS more approachable by presenting it from a new perspective of latent variable models and the so-called bits-back trick.
We guide the reader step by step to a complete implementation of ANS in the Python programming language, which we then generalize for more advanced use cases.
We also present and empirically evaluate an open-source library of various entropy coders designed for both research and production use.%
\footnote{Software library at \url{https://bamler-lab.github.io/constriction} and analyzed in Section~\ref{sec:practice}.}
Related teaching videos and problem sets are available online.%
\footnote{Course modules 5, 6, and~7 at \url{https://robamler.github.io/teaching/compress21}}
\end{abstract}

\section{Introduction}

Effective data compression is becoming increasingly important.
Digitization in business and science raises the demand for data storage, and the proliferation of remote working arrangements makes high-quality videoconferencing indispensable.
Recently, novel compression methods that employ probabilistic machine-learning models have been shown to outperform more traditional compression methods for images and videos~\citep{balle2018variational,minnen2018joint,yang2020improving,agustsson2020scale,yang2021insights}.
Machine learning provides new methods for the \emph{declarative} task of expressing complex probabilistic models, which are essential for data compression (see Section~\ref{sec:fundamentals}).
However, many researchers who come to data compression from a machine learning background struggle to understand the more \emph{procedural} (i.e., algorithmic) parts of a compression pipeline.

This paper discusses an integral algorithmic part of lossless and lossy data compression called entropy coding.
We focus on the Asymmetric Numeral Systems (ANS) entropy coder~\citep{duda2015use}, which is the method of choice in many recent proposals of machine-learning based compression methods \citep{townsend2019practical,kingma2019bit,yang2020improving,theis2021importance}.
We intend to make the internals of ANS more approachable to statisticians and machine learning researchers by presenting them from a new perspective based on latent variable models and the so-called bits-back trick~\citep{wallace1990classification, hinton1993keeping}, which we explain below.
We show that this new perspective allows us to come up with novel variants of ANS that enable research on new compression methods which combine declarative with procedural tasks.

This paper is written in an educational style that favors real code examples over pseudocode, culminating in a working demo implementation of ANS in the Python programming language (Listing~\ref{listing:ans} on page~\pageref{listing:ans}).
The paper is structured as follows:
Section~\ref{sec:background} reviews the relevant information theory and introduces the concept of stream codes.
Section~\ref{sec:pns-to-ans} presents our statistical perspective on the ANS algorithm and guides the reader to a full implementation.
In Section~\ref{sec:beyond}, we show that understanding the internals of ANS allows us to create new variations on it for special use cases.
Section~\ref{sec:practice} presents and empirically evaluates an open-source compression library that is more run-time efficient and easier to use than our demo implementation in Python.
We provide concluding remarks in Section~\ref{sec:conclusion}.

\section{Background}
\label{sec:background}

In Subsection~\ref{sec:fundamentals}, we briefly review the relevant theory of lossless compression.
In Subsection~\ref{sec:symbol-vs-stream}, we summarize how the Asymmetric Numeral Systems (ANS) method differs from other stream codes (like Arithmetic Coding or Range Coding) and from symbol codes (like Huffman Coding).

\subsection{Fundamentals of Lossless Compression}
\label{sec:fundamentals}

We formalize the problem of lossless compression and state the theoretical bounds from the source coding theorem.
Readers who are already familiar with source coding may opt to skip this section.

\paragraph{Problem Statement.}
A lossless compression algorithm (or ``code'')~$C$ maps any message~$\vx$ from some discrete messages space~$\vXX$ to a bit string $C(\vx)$.
We want to make these bit strings~$C(\vx)$ as short as possible while still keeping the mapping~$C$ injective, so that a decoder can reconstruct the message~$\vx$ by inverting~$C$.
We refer to the length of the bit string $C(\vx)$ as the bitrate $R_C(\vx)$.
Since $R_C(\vx)$ may depend on~$\vx$, a common (albeit not the only reasonable) goal in the lossless compression literature is to find a code~$C$ that minimizes the \emph{expected} bitrate $\EE_P[R_C(\vX)]$.
Here, $P$~is a probability distribution over~$\vXX$, i.e., a probabilistic model of the data source (called the ``entropy model'' for short).
Further, $\EE_P[\,\cdot\,]$ is the expectation value, and capital letters like~$\vX$ denote random variables.

To minimize $\EE_P[R_C(\vX)]$, a code~$C$ exploits knowledge of~$P$ to map probable messages to short bit strings while mapping less probable messages to longer bit strings (to avoid clashes that would make $C$ non-injective).
Thus, the bitrate $R_C(\vx)$ varies across messages $\vx\in\vXX$.
In order to prevent~$C$ from encoding information about~$\vx$ into the length rather than the contents of $C(\vx)$, we will assume in the following that~$C$ is \emph{uniquely decodable}.
This means that it must be possible to reconstruct a \emph{sequence} of messages from the (delimiter-free) concatenation of their compressed representations.

\paragraph{Theoretical Bounds.}
The source coding theorem~\citep{shannon1948mathematical} makes two fundamental statements---a negative and a positive one---about the expected bitrate, relating it to the entropy $H_P[\vX]:=\EE_P[-\log_2 P(\vX)]$ of the message~$\vX$ under the entropy model~$P$.
The negative part of the theorem states that the expected bitrate of any uniquely decodable lossless code cannot be smaller than the entropy of the message.
The positive part states that, if we allow for up to one bit of overhead over the entropy, then there indeed always exists a uniquely decodable lossless code.
More formally,
\begin{alignat}{2}
    &\text{$\forall$ entropy models $P$, $\forall$ uniquely decodable lossless codes~$C$:}\quad
    &&\EE_P[R_C(\vX)] \geq H_P[\vX] \label{eq:source-coding-theorem-lower};\; \text{and} \\[2pt]
    &\text{$\forall$ entropy models $P$, $\exists$ uniquely decodable lossless code~$C$:}\quad
    &&\EE_P[R_C(\vX)] < H_P[\vX] + 1 \label{eq:source-coding-theorem-upper}.
\end{alignat}

Thus, the expected bitrate of an \emph{optimal} code~$C_P^*$ lies in the half-open interval $[ H_P[\vX], H_P[\vX]+1)$.
A so-called entropy coder takes an entropy model~$P$ as input and constructs a near-optimal code for it.
The model~$P$ and the choice of entropy coder together comprise a compression method on which two parties have to agree before they can meaningfully exchange any compressed data.

\paragraph{Information Content.}
The upper bound in Eq.~\ref{eq:source-coding-theorem-upper} is actually a corollary of a more powerful statement.
Recall that the entropy $H_P[\vX]=\EE_P[-\log_2 P(\vX)]$ is an expectation over the quantity ``$-\log_2 P(\vX)$''.
This quantity is called the information content of~$\vX$.
Eq.~\ref{eq:source-coding-theorem-upper} thus relates two expectations under~$P$, and it turns out that there exists a code (the so-called Shannon Code~\citep{shannon1948mathematical}) where the inequality holds not only in expectation but indeed for all messages individually, i.e.,
\begin{align}
    \!\!\text{$\forall$ entropy models~$P$, $\exists$ uniqly.~dec.~lossless code~$C$: $\forall \vx\!\in\!\vXX\!:\, R_C(\vx) < -\log_2 P(\vX\eq \vx)+1$}
    \label{eq:bound-shannon-code}.
\end{align}
For large message spaces (e.g., the space of all conceivable HD images), $P(\vX\eq \vx)$ is usually very small for any particular~$\vx$, and the ``$+1$'' on the right-hand side of Eq.~\ref{eq:bound-shannon-code} is negligible compared to the information content $-\log_2 P(\vX\eq \vx)$.
In this regime, the information content of a message~$\vx$ is thus a very accurate estimate of its bitrate $R_{C_P^*}(\vx)$ under a (hypothetical) optimal code~$C_P^*$.

\paragraph{Trivial Example: Uniform Entropy Model.}
To obtain some intuition for Eqs.~\ref{eq:source-coding-theorem-lower}-\ref{eq:bound-shannon-code}, we briefly consider a trivial case, from which our discussion of ANS in Section~\ref{sec:pns-to-ans} below will start.
Consider a finite message space~$\vXX$ with a \emph{uniform} entropy model~$P$, i.e., ${P(\vX \eq \vx)} = {1}/{|\vXX|}\;\forall\vx\!\in\!\vXX$.
Thus, each message $\vx\in\vXX$ has information content $-\log_2 {P(\vX \eq \vx)} = {\log_2 |\vXX|}$, and we have $H_P[\vX] = \log_2|\vXX|$.
We can trivially find a uniquely decodable code~$C_\text{unif.}$ that satisfies Eqs.~\ref{eq:source-coding-theorem-upper}-\ref{eq:bound-shannon-code} for this model~$P$:
we simply enumerate all $\vx\in\vXX$ with integer labels from $\{0,\ldots,|\vXX|-1\}$, which we then express in binary.
The binary representation of the largest label $(|\vXX|-1)$ works out to be $\lceil\log_2 |\vXX|\rceil$~bits long where $\lceil\cdot\rceil$~denotes rounding up to the nearest integer.
To ensure unique decodability, we pad shorter bit strings to the same length with leading zeros.
Thus, $R_{C_\text{unif.}}(\vx) = \left\lceil\log_2|\vXX|\right\rceil = \left\lceil -\log_2 P(\vX\eq \vx)\right\rceil$ for all $\vx\in\vXX$, which satisfies Eq.~\ref{eq:bound-shannon-code} and therefore also Eq.~\ref{eq:source-coding-theorem-upper}.
According to Eq.~\ref{eq:source-coding-theorem-lower}, this trivial code $C_\text{unif.}$ has less than one bit of overhead over the optimal code for such a uniform entropy model.

In practice, however, the entropy model~$P$ is rarely a uniform distribution.
For example, in a compression method for natural language, $P$~should assign higher probability to this paragraph than to some random gibberish.
Any such non-uniform~$P$ has entropy $H_P[\vX] < \log_2|\vXX|$ and thus, according to Eq.~\ref{eq:source-coding-theorem-upper}, there exists a better code for~$P$ than the trivial $C_\text{unif.}$.
In principle, one could find an \emph{optimal} code for any entropy model~$P$ (over a finite~$\vXX$) via the famous Huffman algorithm~\citep{huffman1952method} (knowledge of this algorithm is not required to understand the present paper).
However, directly applying the Huffman algorithm to entire messages would usually be prohibitively computationally expensive.
The next section provides an overview of computationally feasible alternatives.

\subsection{Symbol Codes vs.~Stream Codes}
\label{sec:symbol-vs-stream}

Practical lossless entropy coding methods trade off some compression performance for better runtime and memory efficiency.
There are two common approaches: symbol codes and stream codes~\citep{mackay2003information}.
Both assume that a message $\vx\equiv (x_1,\ldots,x_k)$ is a sequence (with length~$k$) of symbols $x_i\in \XX\; \forall i\in\{1,\ldots,k\}$, with some discrete set (``alphabet'')~$\XX$.
Further, both approaches assume that the entropy model~$P(\vX)=\prod_{i=1}^k P_i(X_i)$ is a product of per-symbol entropy models~$P_i$.%
\footnote{The $P_i$ may form an autoregressive model, i.e., each $P_i$ may be conditioned on all $X_j$ with $j<i$, so that the factorization of $P(\vX)=\prod_i P_i(X_i)$ does not pose any formal restriction on~$P$.}
Thus, the entropy of the message~$\vX$ is the sum of all symbol entropies: $H_P[\vX] = \sum_{i=1}^k H_{P_i}[X_i]$.

Constructing an \emph{optimal} code over the message space~$\vXX = \XX^k$ by Huffman coding would require $O(|\XX|^k)$ runtime (and memory) since Huffman coding builds a tree over all $\vx\in\vXX$.
Both symbol and stream codes reduce the runtime complexity to $O(k)$, but they differ in their precise practical run-time cost and, more fundamentally, in their overhead over the minimal expected bitrate (Eq.~\ref{eq:source-coding-theorem-lower}).

\begin{figure}%
    \centering%
    \includegraphics{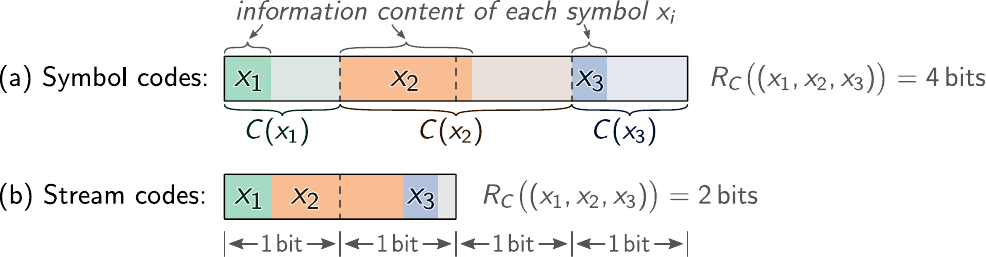}%
    \caption{%
        Comparison between (a)~symbol codes (here: Shannon Coding) and (b)~stream codes (here: Arithmetic Coding).
        Symbol codes assign an integer number of bits to each symbol~$x_i$, which leads to a significant overhead in bitrate if the information content per symbol ($-\log_2 P(X_i \eq x_i)$) is low.
    }%
    \label{fig:symbol-vs-stream}%
\end{figure}
  
\paragraph{Symbol Codes.}
Symbol codes are illustrated in Figure~\ref{fig:symbol-vs-stream}~(a).
They simply apply a uniquely decodable code (e.g., a Huffman code) independently to each symbol~$x_i$ of the message, and then concatenate the resulting bit strings.
This makes the runtime linear in~$k$ but it increases the bitrate.
Even if we use an optimal code~$C_{P_i}^*$ for each symbol~$x_i$, the expected bitrate of each symbol can exceed the symbol's entropy $H_{P_i}[X_i]$ by up to one bit (see Eq.~\ref{eq:source-coding-theorem-upper}).
Since this overhead now applies \emph{per symbol}, the overhead in expected bitrate for the entire message now grows linearly in~$k$.

Symbol codes are used in many classical compression methods such as \texttt{gzip}~\citep{deutsch1996rfc1952} and~\texttt{png}~\citep{boutell1997png}.
Here, the entropy per symbol is large enough that an overhead of up to one bit per symbol can be tolerated.
By contrast, recently proposed machine-learning based compression methods \citep{balle2017end,balle2018variational,yang2020variational,yang2020improving} often spread the total entropy of the message over a larger number~$k$ of symbols, resulting in much lower entropy per symbol.
For example, with about $0.3$~bits of entropy per symbol, a symbol code (which needs at least one full bit per symbol) would have an overhead of more than $200\%$, thus rendering the method useless.

\paragraph{Stream Codes.}
Stream codes reduce the overhead of symbol codes without sacrificing computational efficiency by \emph{amortizing} over symbols.
Similar to symbol codes, stream codes also encode and decode one symbol at a time, and their runtime cost is linear in the message length~$k$.
However, stream codes do not directly map each individual symbol to a bit string.
Instead, for most encoded symbols, a stream code does not immediately emit any compressed bits but instead merely updates some internal encoder state that accumulates information.
Only every once in a while (when the internal state overflows), does the encoder emit some compressed bits that ``pack'' the information content of several symbols more tightly together than in a symbol code (see illustration in Figure~\ref{fig:symbol-vs-stream}~(b)).

The most well-known stream codes are Arithmetic Coding~(AC)~\citep{rissanen1979arithmetic, pasco1976source} and its more efficient variant Range Coding (RC)~\citep{martin1979range}.
This paper focuses on a more recently proposed (and currently less known) stream code called Asymmetric Numeral Systems~(ANS) \citep{duda2015use}.
In practice, all three of these stream codes usually have negligible overhead over the fundamental lower bound in Eq.~\ref{eq:source-coding-theorem-lower} (see Section~\ref{sec:benchmarks} for empirical results).
From a user's point of view, the main difference between these stream codes is that AC and RC operate as a queue (``first in first out'', i.e., symbols are decoded in the same order in which they were encoded) while ANS operates as a stack (``last in first out'').
A queue simplifies compression with autoregressive models while a stack simplifies compression with latent variable models and is generally the simpler data structure for more involved use cases (since reads and writes happen only at a single position).

Further, ANS provides some more subtle advantages over AC and~RC:
\begin{itemize}
\item
    ANS is better suited than AC and RC for advanced compression techniques such as bits-back coding~\citep{wallace1990classification, hinton1993keeping} because (i)~AC and RC treat encoding and decoding asymmetrically with different internal coder states that satisfy different invariants, which complicates switching between encoding and decoding;
    and (ii)~while encoding with RC is, of course, injective, it is difficult to make it surjective (and thus to make decoding arbitrary bit strings injective).
    ANS has neither of these two issues.
\item
    While the main idea of AC and RC is simple, details are somewhat involved due to a number of edge cases.
    By contrast, a complete ANS coder can be implemented in just a few lines of code (see Listing~\ref{listing:ans} in Section~\ref{sec:streaming-ans} below).
    Further, this paper provides a new perspective on ANS that conceptually splits the algorithm into three parts.
    Combining these parts in new ways allows us to design new variants of ANS for special use cases (see Section~\ref{sec:continuity}).
\item
    As a minor point, decoding with AC and RC is somewhat slow because it updates an internal state that is larger than that of the encoder, and because it involves integer division (which is slow on standard hardware~\citep{fog2021instruction}).
    By contrast, decoding with ANS is a very simple operation (assuming that evaluating the model~$P_i$ is cheap, as in a small lookup table).
\end{itemize}

We present the basic ANS algorithm in Section~\ref{sec:pns-to-ans}, discuss possible extensions to it in Section~\ref{sec:beyond}, and empirically compare its compression performance and runtime efficiency to AC and RC in Section~\ref{sec:practice}.

\section{From Positional Numeral Systems to Asymmetric Numeral Systems}
\label{sec:pns-to-ans}

This section guides the reader towards a full implementation of an Asymmetric Numeral Systems (ANS) entropy coder.
We start with a trivial coder for sequences of uniformly distributed symbols (Subsection~\ref{sec:pns}), generalize it to arbitrary entropy models (Subsection~\ref{sec:bits-back}), and finally make it computationally efficient (Subsection~\ref{sec:streaming-ans}).
Our presentation of ANS differs from the original proposal~\citep{duda2015use} in that it builds on latent variable models and the so-called bits-back trick, which we hope makes it more approachable to statisticians and machine learning researchers.

\subsection{A Simple Stream Code: Positional Numeral Systems}
\label{sec:pns}

We introduce our first example of a stream code: a simple code that provides optimal compression performance for sequences of \emph{uniformly} distributed symbols.
This code will serve as a fundamental building block in our implementation of Asymmetric Numeral Systems in Sections~\ref{sec:bits-back}-\ref{sec:streaming-ans} below.

As shown in Section~\ref{sec:fundamentals}, we can obtain an optimal lossless compression code $C_\text{unif.}$ for a uniform entropy model over a finite message space~$\vXX$ by using a bijective mapping from $\vXX$ to $\{0, \ldots, |\vXX| - 1\}$ and expressing the resulting integer in binary.
Such a bijective mapping can be done efficiently if messages are sequences $(x_1,\ldots,x_k)$ of symbols~$x_i\in\XX$ from a finite alphabet~$\XX$:
we simply interpret the symbols as the digits of a number in the positional numeral system of base~$|\XX|$.
For example, if the alphabet size is $|\XX| = 10$ then we can assume, without loss of generality, that $\XX = \{0, 1, \ldots, 9\}$.
To encode, e.g., the message $\vx = (3,6,5)$ we thus simply parse it as the decimal number~$365$, which we then express in binary: $365 = {(101101101)}_2$.
We defer the issue of unique decodability to Section~\ref{sec:streaming-ans}.
Despite its simplicity, this code has three important properties, which we highlight next.

\paragraph{Property 1: ``Stack'' Semantics.}
Encoding and decoding traverse symbols in a message in opposite order.
For encoding, the simplest way to parse a sequence of digits $x_i\in\XX$ into a number in the positional numeral system of base~$|\XX|$ goes as follows:
initialize a variable with zero;
then, for each digit, multiply the variable with~$|\XX|$ and add the digit.
For example, if $|\XX| = 10$ and $\vx=(3,6,5)$:

\begin{tabular}{@{}b{4.7cm}l@{}}
    Parsing the decimal number $365$:\vspace{-1.9pt}&
    \includegraphics{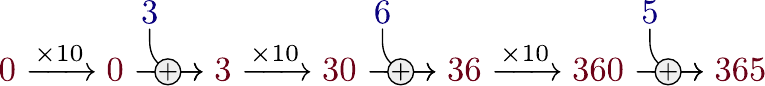}
\end{tabular}

Encoding thus reads the digits of the decimal number $365$ from left to right, i.e., from most to least significant digit.
By contrast, decoding the number back into its digits is easier done in reverse order:
keep dividing the number by the base~$|\XX|$, rounding down and recording the remainders as the digits:

\begin{tabular}{@{}b{7cm}l@{}}
    Generating the digits of the decimal number $365$: \vspace{27.6pt} &
    \includegraphics{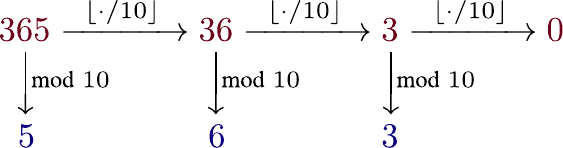}
\end{tabular}

\begin{listing}[t]
\begin{minted}[linenos,frame=lines,escapeinside=@@]{python}
class UniformCoder: # optimal code for uniformly distributed symbols
    def __init__(self, number=0):   # constructor with an optional argument
        self.number = number

    def push(self, symbol, base):   # Encodes a @$\mathtt{symbol} \in\{0, \ldots, \mathtt{base}-1\}$@.
        self.number = self.number * base + symbol

    def pop(self, base):            # Decodes a @$\mathtt{symbol} \in\{0, \ldots, \mathtt{base}-1\}$@.
        symbol = self.number %
        self.number //= base        # “@{\normalfont\texttt{//}}@” denotes integer division.
        return symbol
\end{minted}
\vspace{-4pt}
\caption{
    Python implementation of a stack-based optimal encoder/decoder for sequences of uniformly distributed symbols, as discussed in Section~\ref{sec:pns};
    for usage examples, see Listings~\ref{listing:pns-example} and~\ref{listing:pns-heterogeneous}.
}
\label{listing:pns}
\end{listing}

\begin{listing}[t]
\begin{minted}[linenos,frame=lines,escapeinside=||]{python}
coder = UniformCoder() # Defined in Listing |\ref*{listing:pns}|.

# Encode the message |$\vx=(3,6,5)$|:
coder.push(3, base=10) # |{\normalfont\texttt{base=10}}| means that the alphabet is |$\XX=\{0, \ldots, 9\}$|.
coder.push(6, base=10)
coder.push(5, base=10)
print(f"Encoded number: {coder.number}")  # Prints: “Encoded number: 365”
print(f"In binary: {coder.number:b}")     # Prints: “In binary: 101101101”

# Decode the encoded symbols (in reverse order):
print(f"Decoded '{coder.pop(base=10)}'.") # Prints: “Decoded '5'.”
print(f"Decoded '{coder.pop(base=10)}'.") # Prints: “Decoded '6'.”
print(f"Decoded '{coder.pop(base=10)}'.") # Prints: “Decoded '3'.”
\end{minted}
\vspace{-4pt}
\caption{
    Encoding and decoding the message $\vx=(3,6,5)$ using our \texttt{UniformCoder} from Listing~\ref{listing:pns};
    digits are encoded from left to right but decoded from right to left, i.e., the coder has \emph{stack} semantics.}
\label{listing:pns-example}
\end{listing}

Thus, the simplest implementation of $C_\text{unif.}$ has ``stack'' semantics (last-in-first-out): conceptually, encoding \emph{pushes} information onto the top of a stack of compressed data, and decoding \emph{pops} the latest encoded information off the top of the stack.
Listing~\ref{listing:pns} implements these \texttt{push} and \texttt{pop} operations in Python as methods on a class \texttt{UniformCoder}.
A simple usage example is given in Listing~\ref{listing:pns-example} (all code examples in this paper are also available online%
\footnote{\url{https://github.com/bamler-lab/understanding-ans/blob/main/code-examples.ipynb}}
as an executable jupyter notebook.)

\paragraph{Property 2: Amortization.}
Before we explore more advanced usages of our \texttt{UniformCoder} from Listing~\ref{listing:pns}, we illustrate how this stream code amortizes compressed bits over multiple symbols (see Section~\ref{sec:symbol-vs-stream}).
Consider again encoding the message $\vx=(3,6,5)$ with the alphabet $\XX=\{0,\ldots, 9\}$.
Modifying one of the symbols in this message obviously changes the compressed bit string, e.g.,
\begin{align}
    &\begin{array}{r@{$\;=\;$}c@{$\;\!$}c@{$\;\!$}c@{$\;\!$}c@{$\;\!$}c@{$\;\!$}c@{$\;\!$}c@{$\;\!$}c@{$\;\!$}c@{$\;\!$}l}
        C_\text{unif.}\big((3,6,5)\big) & 1&0&1&1&0&1&1&{\color{red!70!black} 0}&1&; \\[3pt]
        C_\text{unif.}\big((3,6,\underline{6})\big) & 1&0&1&1&0&1&1&\underline{\color{red!70!black} 1}&\underline{0}&; \\[3pt]
        C_\text{unif.}\big((3,\underline{7}, 5)\big) & 1&0&1&1&\underline{1}&\underline{0}&1&\underline{\color{red!70!black} 1} &1&.
    \end{array}
\end{align}
Here, underlined numbers highlight differences to $C_\text{unif.}\big((3,6,5)\big)$ (first line).
Note that the penultimate bit (red) changes from ``0'' to ``1'' when we change the third symbol from $x_3=5$ to $x_3=6$ (second line), and also when we change the second symbol from $x_2=6$ to $x_2=7$ (third line).
Thus, this bit depends on both $x_2$ and~$x_3$.
This is a crucial difference to symbol codes (e.g., Huffman Codes), which attribute each bit in the compressed bit string to exactly one symbol only (see Figure~\ref{fig:symbol-vs-stream}~(a)).
Amortization is what allows stream codes to reach lower expected bitrates than symbol codes.

\begin{listing}[t]
\begin{minted}[linenos,frame=lines,escapeinside=||]{python}
coder = UniformCoder() # Defined in Listing |\ref*{listing:pns}|.

# Encode a sequence of symbols from alphabets of varying sizes:
coder.push( 3, base=10) # |{\normalfont\texttt{base=10}}| means that the alphabet is |$\{0, \ldots, 9\}$|.
coder.push( 6, base=10)
coder.push(12, base=15) # Setting |{\normalfont\texttt{base=15}}| switches to alphabet |$\{0, \ldots, 14\}$|.
coder.push( 4, base=15)
print(f"Binary: {coder.number:b}") # Prints: “Binary: 10000001011100”

# For decoding, use the same sequence of bases but in *reverse* order:
print(f"Decoded '{coder.pop(base=15)}'.") # Prints: “Decoded '4'.”
print(f"Decoded '{coder.pop(base=15)}'.") # Prints: “Decoded '12'.”
print(f"Decoded '{coder.pop(base=10)}'.") # Prints: “Decoded '6'.”
print(f"Decoded '{coder.pop(base=10)}'.") # Prints: “Decoded '3'.”
\end{minted}
\vspace{-4pt}
\caption{
    Encoding and decoding the message $\vx=(3,6,12,4)$ where the first two symbols come from the alphabet $\{0,\ldots,9\}$ and the last two symbols come from the alphabet $\{0,\ldots,14\}$.
}
\label{listing:pns-heterogeneous}
\end{listing}

\paragraph{Property 3: Varying Alphabet Sizes.}
Our implementation of ANS in Sections~\ref{sec:bits-back}-\ref{sec:streaming-ans} below will exploit an additional property of the \texttt{UniformCoder} from Listing~\ref{listing:pns}:
each symbol~$x_i$ of the message $\vx=(x_1,\ldots,x_k)$ may come from an individual alphabet~$\XX_i$, with individual alphabet size~$|\XX_i|$.
We just have to set the \texttt{base} when encoding or decoding each symbol~$x_i$ to the corresponding alphabet size~$|\XX_i|$.
For example, in Listing~\ref{listing:pns-heterogeneous} we encode and decode the example message $\vx\eq(3,6,12,4)$ using the alphabet $\XX_1\eq\XX_2\eq\{0,\ldots,9\}$ for the first two symbols and the alphabet $\XX_3\eq\XX_4\eq\{0,\ldots,14\}$ for the last two symbols.
The example works because the method \texttt{pop} of \texttt{UniformCoder} (Listing~\ref{listing:pns}) performs the exact inverse of \texttt{push}:
calling \texttt{push(symbol, base)} followed by \texttt{pop(base)} with identical \texttt{base} (and with $\texttt{symbol} \in \{0, \ldots, \texttt{base}-1\}$) restores the coder's internal state, regardless of its history.
This generalized use of a \texttt{UniformCoder} still maps bijectively between the message space (which is now $\vXX= \XX_1\times \XX_2\times \ldots \times\XX_k$) and the integers $\{0,\ldots,|\vXX| -1\}$.
It therefore still provides optimal compression performance assuming that all symbols are uniformly distributed over their respective alphabets.
In the next section, we lift this restriction on uniform entropy models.

\subsection{Latent Variable Models and the Bits-Back Trick}
\label{sec:bits-back}

In this section, we use the \texttt{UniformCoder} introduced in the last section (Listing~\ref{listing:pns}) as a building block for a more general stream code that is no longer limited to uniform entropy models.
This leads us to a precursor of ANS that provides very close to optimal compression performance for arbitrary entropy models, but that is too slow for practical use.
We will speed it up in Section~\ref{sec:streaming-ans} below.

\paragraph{Approximated Entropy Model.}
The \texttt{UniformCoder} from Listing~\ref{listing:pns} minimizes the expected bitrate $\EE_P[R_{C_\text{unif.}}(\vX)]$ only if the entropy model~$P$ is a uniform distribution.
We now consider the more general case where $P(\vX)=\prod_{i=1}^k P_i(X_i)$ still factorizes over all symbols, but each $P_i$ is now an arbitrary distribution over a finite alphabet~$\XX_i$.
The ANS algorithm approximates each~$P_i$ with a distribution~$Q_i$ that represents the probability $Q_i(X_i \eq x_i)$ of each symbol~$x_i$ in fixed point precision,
\begin{align}\label{eq:approximatep}
    Q_i(X_i \eq x_i) = \frac{m_i(x_i)}{n}
    \quad\forall x_i\in\XX_i,
    \qquad\text{where $n=2^\mathtt{precision}$.}
\end{align}
Here, $\mathtt{precision}$ is a positive integer (typically $\lesssim 32$) that controls a trade-off between computational cost and accuracy of the approximation.
Further, $m_i(x_i)$ for ${x_i\in\XX_i}$ are integers that should ideally be chosen by minimizing the Kullback-Leibler (KL) divergence $D_\text{KL}(P_i \,||\, Q_i)$, which measures the overhead (in expected bitrate) due to approximation errors.
In practice, however, exact minimization of this KL-divergence is rarely necessary as simpler heuristics (e.g., by lazily rounding the cumulative distribution function) usually already lead to a negligible overhead for reasonable \texttt{precision}s.
Note however, that setting $m_i(x_i)=0$ for any ${x_i\in\XX_i}$ makes it impossible to encode~$x_i$ with ANS, and that correctness of the ANS algorithm relies on $Q_i$ being properly normalized, i.e., $\sum_{x_i\in\XX_i} m_i(x_i)=n$.

ANS uses the approximated entropy models~$Q_i$ for its encoding and decoding operations.
While the operations turn out to be very compact, it is often somewhat puzzling to users why they work, why they provide near-optimal compression performance, and how they could be extended to satisfy potential additional constraints for special use cases.
In the following, we aim to make ANS more approachable by proposing a new perspective on it that expresses~$Q_i$ as a latent variable model.

\begin{figure}%
    \centering%
    \parbox{71mm}{%
        \centering%
        \includegraphics{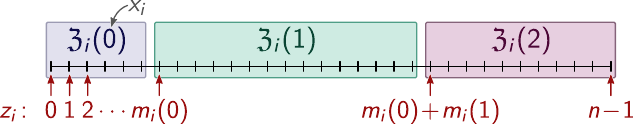}%
        \caption{%
            Partitioning of the range $\{0,\ldots,{n\!-\!1}\}$ into disjoint subranges $\ZZ_i(x_i)$, $x_i\in\XX_i$, see Eq.~\ref{eq:def-subrange}.
            The size of each subrange, $|\ZZ_i(x_i)|=m_i(x_i)$, is proportional to the probability $Q_i(X_i\eq x_i)$ (Eq.~\ref{eq:approximatep}).}%
        \label{fig:subranges}%
    }%
    \hspace{\dimexpr\textwidth-71mm-63mm\relax}%
    \begin{minipage}{63mm}%
        \centering%
        \includegraphics{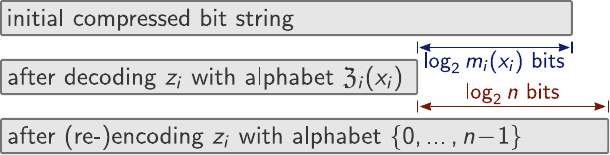}%
        \caption{%
            Encoding a symbol~$x_i$ with ANS using the bits-back trick.
            The net bitrate is $\log_2 n - \log_2 m_i(x_i) = -\log_2 Q_i(X_i\eq x_i)$.
        }%
        \label{fig:bitsback}%
    \end{minipage}%
\end{figure}

\paragraph{Latent Variable Model.}
Since the integers $m_i(x_i)$ in Eq.~\ref{eq:approximatep} sum to~$n$ for $x_i\in\XX_i$, they define a partitioning of the integer range $\{0,\ldots,n-1\}$ into $|\XX_i|$ disjoint subranges, see Figure~\ref{fig:subranges}:
for each symbol $x_i\in\XX_i$, we define a subrange $\ZZ_i(x_i) \subseteq \{0,\ldots,n-1\}$ of size $|\ZZ_i(x_i)|=m_i(x_i)$ as follows,
\begin{align}\label{eq:def-subrange}
    \textstyle \ZZ_i(x_i) := \left\{\sum_{x_i'<x_i} m_i(x_i'), \;\;\ldots,\;\; \left(\sum_{x_i'\leq x_i} m_i(x_i')\right)-1 \right\},
\end{align}
where we assumed some ordering on the alphabet~$\XX_i$ (to define what ``$x_i' \!<\! x_i$'' means).
By construction, the subranges $\ZZ_i(x_i)$ are pairwise disjoint for different~$x_i$ and they cover the entire range $\{0,\ldots,n-1\}$ (see Figure~\ref{fig:subranges}).
Imagine now we want to draw a random sample~$x_i$ from~$Q_i$.
A simple way to do this is to draw an auxiliary integer $z_i$ from a uniform distribution over $\{0,\ldots,n-1\}$, and then identify the unique symbol~$x_i\in\XX_i$ that satisfies $z_i\in\ZZ_i(x_i)$.
This procedure describes the so-called generative process of a latent variable model $Q_i(X_i,Z_i) = Q(Z_i) \, Q_i(X_i\,|\,Z_i)$ where
\begin{align}\label{eq:prior-likelihood}
    Q(Z_i \eq z_i) &= \frac1n \;\;\forall z_i\in\{0,\ldots,n-1\}
    \,\quad\text{and}\,\quad
    Q_i(X_i\eq x_i \,|\, Z_i\eq z_i) = \begin{cases}
        1 & \!\!\text{if $z_i \in \ZZ_i(x_i)$;} \\ 
        0 & \!\!\text{else.}
    \end{cases}
\end{align}
Using $|\ZZ_i(x_i)| = m_i(x_i)$, one can easily verify that the approximated entropy model from Eq.~\ref{eq:approximatep} is the marginal distribution of the latent variable model from Eq.~\ref{eq:prior-likelihood}, i.e., our use of the same name $Q_i$ for both models is justified since we indeed have $Q_i(X_i \eq x_i) = \sum_{z_i=0}^{n-1} Q_i(X_i \eq x_i,Z_i \eq z_i)\; \forall x_i\in\XX_i$.

The uniform prior distribution $Q(Z_i)$ in Eq.~\ref{eq:prior-likelihood} suggests that we could build a stream code by utilizing our \texttt{UniformCoder} from Listing~\ref{listing:pns}.
We begin with a naive approach upon which we then improve:
to encode a message $\vx=(x_1,\ldots,x_k)$, we simply identify each symbol~$x_i$ by some~$z_i$ that we may choose arbitrarily from $\ZZ_i(x_i)$, and we encode the resulting sequence $(z_1,\ldots,z_k)$ using a \texttt{UniformCoder} with alphabet $\{0,\ldots,n-1\}$.
On the decoder side, we decode $(z_1,\ldots,z_k)$ and we recover the message~$\vx$ by identifying, for each~$i$, the unique symbol~$x_i$ that satisfies $z_i\in\ZZ_i(x_i)$.

\paragraph{The Bits-Back Trick.}
Unfortunately, the above naive approach suffers from poor compression performance.
Encoding~$z_i$ with the uniform prior model~$Q(Z_i)$ increases the (amortized) bitrate by $z_i$'s information content, $-\log_2 Q(Z_i\eq z_i) = -\log_2 \frac1n= \texttt{precision}$.
By contrast, an optimal code would spend only $-\log_2 Q_i(X_i\eq x_i) = \texttt{precision} - \log_2 m_i(x_i)$~bits on symbol~$x_i$ (see Eq.~\ref{eq:approximatep}).
Thus, our naive approach has an overhead of $\log_2 m_i(x_i)$~bits for each symbol~$x_i$.
The overhead arises because the encoder can choose~$z_i$ arbitrarily among any element of $\ZZ_i(x_i)$, and this choice injects some information into the compressed bit string that is discarded on the decoder side.
This suggests a simple solution, called the bits-back trick~\citep{wallace1990classification, hinton1993keeping}: 
we should somehow utilize the $\log_2 m_i(x_i)$ bits of information contained in our choice of $z_i\in \ZZ_i(x_i)$, e.g., by communicating through it some part of the previously encoded symbols.

\begin{listing}[t]
\begin{minted}[linenos,frame=lines,escapeinside=||]{python}
class SlowAnsCoder: # Has near-optimal bitrates but high runtime cost.
    def __init__(self, precision, compressed=0):
        self.n = 2**precision  # See |Eq. \ref*{eq:approximatep}| (“|{\normalfont\texttt{**}}|” denotes exponentiation).
        self.stack = UniformCoder(compressed) # Defined in Listing |\ref*{listing:pns}|.|\label{line:slow_ans_def_stack}|

    def push(self, symbol, m): # Encodes one symbol.
        z = self.stack.pop(base=m[symbol]) + sum(m[0:symbol])|\label{line:slow_ans_push1}|
        self.stack.push(z, base=self.n)|\label{line:slow_ans_push2}|

    def pop(self, m):          # Decodes one symbol.
        z = self.stack.pop(base=self.n)|\label{line:slow_ans_pop1}|
        # Find the unique |$\mathtt{symbol}$| that satisfies |$\mathtt{z} \in \ZZ_i(\mathtt{symbol})$| (real deploy-
        # ments should use a more efficient method than linear search):
        for symbol, m_symbol in enumerate(m):|\label{line:slow_ans_find_symbol_begin}|
            if z >= m_symbol:
                z -= m_symbol
            else:
                break|\label{line:slow_ans_find_symbol_end}|
        self.stack.push(z, base=m_symbol)|\label{line:slow_ans_pop2}|
        return symbol

    def get_compressed(self):
        return self.stack.number
\end{minted}
\vspace{-4pt}
\caption{
    A coder with near-optimal compression performance for arbitrary entropy models due to the bits-back trick (Section~\ref{sec:bits-back}), but with poor runtime performance.
    Usage example in Listing~\ref{listing:ans_usage}.
}
\label{listing:slowans}
\end{listing}

Listing~\ref{listing:slowans} implements this bits-back trick in a class \texttt{SlowAnsCoder} (a precursor to an \texttt{AnsCoder} that we will introduce in the next section).
Like our \texttt{UniformCoder}, a \texttt{SlowAnsCoder} operates as a stack.
The method \texttt{push} for encoding accepts a \texttt{symbol}~$x_i$ from the (implied) alphabet $\XX_i = {\{0,\ldots, \texttt{len(m)}-1\}}$, where the additional method argument ``\texttt{m}'' is a list of all the integers $m_i(x_i')$ for $x_i'\in\XX_i$, i.e., \texttt{m}~specifies a model~$Q_i$ (Eq.~\ref{eq:approximatep}), and \texttt{m[symbol]} is $m_i(x_i)$.
The method body is very brief:
Line~\ref*{line:slow_ans_push1} in Listing~\ref{listing:slowans} picks a $z_i\in\ZZ_i(x_i)$ and Line~\ref*{line:slow_ans_push2} encodes~$z_i$ onto a \texttt{stack} (which is a \texttt{UniformCoder}, see Line~\ref*{line:slow_ans_def_stack}) using a uniform entropy model over $\{0,\ldots,n-1\}$.
The interesting part is how we pick~$z_i$ in Line~\ref*{line:slow_ans_push1}:
we \emph{decode}~$z_i$ from \texttt{stack} using the alphabet~$\ZZ_i(x_i)$.%
\footnote{Technically, this is done by decoding with the shifted alphabet $\{0,\ldots,m_i(x_i)-1\}$ and then shifting the decoded value back by adding \texttt{sum(m[0:symbol])}, which is Python notation for $\sum_{x_i'<x_i} m_i(x_i')$.}
Note that, at this point, we haven't actually encoded~$z_i$ yet;
we just decode it from whatever data has accumulated on \texttt{stack} from previously encoded symbols (if any).
This gives us no control over the value of~$z_i$ except that it is guaranteed to lie within $\ZZ_i(x_i)$, which is all we need.
Crucially, decoding~$z_i$ consumes data from \texttt{stack} (see Figure~\ref{fig:bitsback}), thus reducing the (amortized) bitrate as we analyze below.

The method \texttt{pop} for decoding inverts each step of \texttt{push}, and it does so in reverse order because we are using a \texttt{stack}.
Line~\ref*{line:slow_ans_pop1} in Listing~\ref{listing:slowans} decodes~$z_i$ using the alphabet $\{0,\ldots,n-1\}$ (thus inverting Line~\ref*{line:slow_ans_push2}) and Line~\ref*{line:slow_ans_pop2} encodes~$z_i$ using the alphabet $\ZZ_i(x_i)$ (thus inverting Line~\ref*{line:slow_ans_push1}).
The latter step recovers the information that we communicate trough our choice of $z_i\in\ZZ_i(x_i)$ in \texttt{push}.
The method \texttt{pop} may appear more complicated than \texttt{push} because it has to find the unique symbol~$x_i$ that satisfies $z_i\in\ZZ_i(x_i)$, which is implemented here---for demonstration purpose---with a simple linear search (Lines~\ref*{line:slow_ans_find_symbol_begin}-\ref*{line:slow_ans_find_symbol_end}).
This search simultaneously subtracts $\sum_{x_i'<x_i}m_i(x_i')$ from~$z_i$ before encoding it, which inverts the part of Line~\ref*{line:slow_ans_push1} that adds \texttt{sum(m[0:symbol])} to the value decoded from \texttt{stack}.

\begin{listing}[t]
\begin{minted}[linenos,frame=lines,escapeinside=||]{python}
# Specify an approximated entropy model via |$\mathtt{precision}$| and |$m_i(x_i)$| from Eq. |\ref*{eq:approximatep}|:
precision = 4 # For demonstration; deployments should use higher |$\mathtt{precision}$|.|\label{line:set_precision}|
m = [7, 3, 6] # Sets |$Q_i(X_i\eq 0) = \frac{7}{2^4}$|, |$Q_i(X_i\eq 1) = \frac{3}{2^4}$|, and |$Q_i(X_i\eq 2) = \frac{6}{2^4}$|.|\label{line:set_m}|

# Encode a message in |$\mathtt{reversed}$| order so we can decode it in forward order:
example_message = [2, 0, 2, 1, 0]|\label{line:ans_example_encode_begin}|
encoder = SlowAnsCoder(precision) # Also works with |$\mathtt{AnsCoder}$| (Listing |\ref*{listing:ans}|).
for symbol in reversed(example_message):|\label{line:ans_example_encoder_loop}|
    encoder.push(symbol, m) # We could use a different |$\mathtt{m}$| for each |$\mathtt{symbol}$|.
compressed = encoder.get_compressed()

# We could actually reuse the |$\mathtt{encoder}$| for decoding, but let's pretend that
# decoding occurs on a different machine that receives only “|$\mathtt{compressed}$|”.
decoder = SlowAnsCoder(precision, compressed)|\label{line:ans_example_decode_begin}|
reconstructed = [decoder.pop(m) for _ in range(5)]|\label{line:ans_example_decoder_loop}|
assert reconstructed == example_message # Verify correctness.|\label{line:ans_example_decode_end}|
\end{minted}
\vspace{-4pt}
\caption{
    Simple usage example for our \texttt{SlowAnsCoder} from Listing~\ref{listing:slowans};
    the same example also works with our more efficient \texttt{AnsCoder} from Listing~\ref{listing:ans}.
    For simplicity, we use the same entropy model~$Q_i$ (Lines~\ref*{line:set_precision}-\ref*{line:set_m}) for all symbols, but we could easily use an individual model for each symbol.
}
\label{listing:ans_usage}
\end{listing}

Listing~\ref{listing:ans_usage} shows a usage example for \texttt{SlowAnsCoder}.
For simplicity, we encode each symbol with the same model~$Q_i$ here, but it would be straight-forward to use an individual model for each symbol.

\begin{listing}[t]
\begin{minted}[linenos,frame=lines,escapeinside=||]{python}
class SlowAnsCoder: # Equivalent to Listing |\ref*{listing:slowans}|, just more self-contained.
    def __init__(self, precision, compressed=0):
        self.n = 2**precision # See |Eq. \ref*{eq:approximatep}| (“**” denotes exponentiation).
        self.compressed = compressed  # (== |$\mathtt{stack}$.$\mathtt{number}$| in |$\mathtt{SlowAnsCoder}$|)

    def push(self, symbol, m):        # Encodes one symbol.
        z = self.compressed %
        self.compressed //= m[symbol] # “//” denotes integer division.|\label{line:slow_ans_inlined_push1b}|
        self.compressed = self.compressed * self.n + z|\label{line:slow_ans_inlined_push2}|

    def pop(self, m):                 # Decodes one symbol.
        z = self.compressed %
        self.compressed //= self.n    # “//” denotes integer division.|\label{line:slow_ans_inlined_pop1b}|
        # Identify |{\normalfont\texttt{symbol}}| and subtract |{\normalfont\texttt{sum(m[0:symbol])}}| from |{\normalfont\texttt{z}}|:
        for symbol, m_symbol in enumerate(m):|\label{line:slow_ans_inlined_find_symbol_begin}|
            if z >= m_symbol:
                z -= m_symbol
            else:
                break # We found the |$\mathtt{symbol}$| that satisfies |$\mathtt{z} \in \ZZ_i(\mathtt{symbol})$|.|\label{line:slow_ans_inlined_find_symbol_end}|
        self.compressed = self.compressed * m_symbol + z|\label{line:slow_ans_inlined_pop2}|
        return symbol

    def get_compressed(self):
        return self.compressed
\end{minted}
\vspace{-4pt}
\caption{
    Reimplementation of \texttt{SlowAnsCoder} from Listing~\ref{listing:slowans} where we removed the dependency on \texttt{UniformCoder} (Listing~\ref{listing:pns}) by manually inlining the method calls \texttt{stack.push()} and \texttt{stack.pop()}.
}
\label{listing:slowans_inlined}
\end{listing}

\paragraph{Correctness and Compression Performance.}
Many researchers are initially confused by the bits-back trick.
It may help to separately analyze its \emph{correctness} and its \emph{compression performance}.
\emph{Correctness} means that the method \texttt{pop} is the exact inverse of \texttt{push}, i.e., calling \hbox{\texttt{push(symbol, m)}} followed by \texttt{pop(m)} returns \texttt{symbol} and restores the \texttt{SlowAnsCoder} into its original state (assuming that all method arguments are valid, i.e., $0\leq \mathtt{symbol} < \texttt{len(m)}$ and $\texttt{sum(m)} = n = 2^\texttt{precision}$).
This can easily be verified by following the implementation in Listing~\ref{listing:slowans} step by step.
Crucially, it holds for any original state of the \texttt{SlowAnsCoder}, even if the coder starts out empty (i.e., with ${\texttt{compressed} = 0}$).
Some readers may find it easier to verify this by reference to Listing~\ref{listing:slowans_inlined}, which reimplements the \texttt{SlowAnsCoder} from Listing~\ref{listing:slowans} in a self-contained way, i.e., without explicitly using a \texttt{UniformCoder} from Listing~\ref{listing:pns} because all method calls to \texttt{stack} are manually inlined.

The \emph{compression performance} of a \texttt{SlowAnsCoder} is analyzed more easily in Listing~\ref{listing:slowans}.
Line~\ref*{line:slow_ans_push1} of the method \texttt{push} \emph{decodes}~$z_i$ from \texttt{stack} with a uniform entropy model over $\ZZ_i(x_i)$, i.e., with the model ${Q_i(Z_i\eq z_i \,|\,X_i\eq x_i)} = 1/m_i(x_i)\;\forall z_i\in \ZZ_i(x_i)$.
This reduces the (amortized) bitrate by the information content $-\log_2 Q_i(Z_i\eq z_i\,|\,X_i\eq x_i) = \log_2 m_i(x_i)$~bits, provided that at least this much data is available on \texttt{stack} (which holds for all but the first encoded symbol in a message).
Line~\ref*{line:slow_ans_push2} then \emph{encodes} onto \texttt{stack} with a uniform entropy model over $\{0,\ldots,n-1\}$, which increases the bitrate by $\log_2 n$.
Thus, in total, encoding (\texttt{push}ing) a symbol~$x_i$ contributes $\log_2\frac{n}{m_i(x_i)}$~bits (see Figure~\ref{fig:bitsback}), which is precisely the symbol's information content, $-\log_2 Q_i(X_i\eq x_i)$ (see Eq.~\ref{eq:approximatep}).
Therefore, the \texttt{SlowAnsCoder} implements an \emph{optimal} code for~$Q$, except for the first symbol of the message, which always contributes $\log_2 n = \mathtt{precision}$~bits (this constant overhead is negligible for long messages).
However, the \texttt{SlowAnsCoder} turns out to be slow (as its name suggests), which we address next.

\subsection{Computational Efficiency: The Practical (Streaming) ANS Algorithm}
\label{sec:streaming-ans}

We speed up the \texttt{SlowAnsCoder} introduced in the last section (Listings~\ref{listing:slowans} or~\ref{listing:slowans_inlined}), which finally leads us to the variant of ANS that is commonly used in practice (also called ``streaming ANS'').
While the \texttt{SlowAnsCoder} provides near-optimal compression performance, it is not a very practical algorithm because the runtime cost for encoding a message of length~$k$ scales quadratically rather than linearly in~$k$.
This is because \texttt{SlowAnsCoder} represents the entire compressed bit string as a single integer (\texttt{compressed} in Listing~\ref{listing:slowans_inlined}), which would therefore become extremely large (typically millions of bits long) in practical applications such as image compression.%
\footnote{
    The reason why our \texttt{SlowAnsCoder} works at all is because Python seamlessly switches to a ``big int'' representation for large numbers.
    A similarly naive implementation in C++ or Rust would overflow very quickly.
}
The runtime cost of arithmetic operations involving \texttt{compressed} (Lines~\ref*{line:slow_ans_inlined_push1a}, \ref*{line:slow_ans_inlined_push1b}, and~\ref*{line:slow_ans_inlined_pop2} in Listing~\ref{listing:slowans_inlined}) thus scales linearly with the amount of data that has been compressed so far, leading to an overall cost of $O(k^2)$ for a sequence of $k$~symbols.

To speed up the algorithm, we first observe that not all arithmetic operations on large integers are necessarily slow.
Lines~\ref*{line:slow_ans_inlined_push2}, \ref*{line:slow_ans_inlined_pop1a}, and~\ref*{line:slow_ans_inlined_pop1b} in Listing~\ref{listing:slowans_inlined} perform multiplication, modulo, and integer division between \texttt{compressed} and~$n$.
Since $n=2^\texttt{precision}$ is a power of two (Eq.~\ref{eq:approximatep}), this suggests storing the bits that represent the giant integer \texttt{compressed} in a dynamically sized array (aka ``vector'') of \texttt{precision}-bit chunks (called ``words'' below) so that these arithmetic operations simplify to appending a zero word to the vector, inspecting the last word in the vector, and deleting the last word, respectively (analogous to how ``$\times 10$'', ``$\text{mod}~10$'' and ``$\lfloor \cdot /10\rfloor$'' are trivial operations in the decimal system).
A good vector implementation performs these operations in constant (amortized) runtime.

\begin{figure}%
    \centering%
    \includegraphics[width=\textwidth]{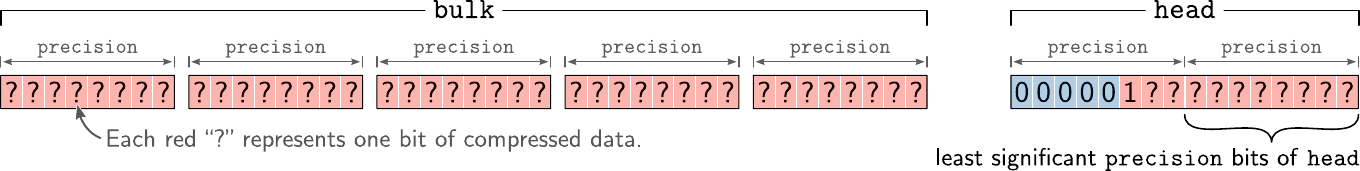}%
    \caption{%
        Streaming ANS with a growable \texttt{bulk} and a finite-capacity \texttt{head}.
        Encoding and decoding operate on \texttt{head}, but if encoding a symbol would overflow \texttt{head} then we first transfer the least sig\-nif\-i\-cant \texttt{precision}~bits of \texttt{head} to \texttt{bulk}.
        More general configurations are possible, see Section~\ref{sec:generalized_streaming}.
    }%
    \label{fig:streaming-ans}%
\end{figure}

Unfortunately, the remaining arithmetic operations (Lines~\ref*{line:slow_ans_inlined_push1a}, \ref*{line:slow_ans_inlined_push1b}, and~\ref*{line:slow_ans_inlined_pop2} in Listing~\ref{listing:slowans_inlined}) cannot be reduced to a constant runtime cost because restricting~$m_i(x_i)$ to a power of two for all $x_i\in\XX_i$ would severely limit the expressivity of the approximated entropy model~$Q$ (Eq.~\ref{eq:approximatep}).
The form of ANS that is used in practice (sometimes called ``streaming ANS'') therefore employs a hybrid approach that splits the compressed bit string into a \texttt{bulk} and a \texttt{head} part (see Figure~\ref{fig:streaming-ans}).
The \texttt{bulk} holds most of the compressed data as a growable vector of fixed-sized words while the \texttt{head} has a relatively small fixed capacity (e.g.,~64~bits).
Most of the time, encoding and decoding operate only on \texttt{head} and thus have a constant runtime due to the bounded size of \texttt{head}.
Only when \texttt{head} overflows or underflows certain thresholds (see below) do we transfer some data between \texttt{head} and the end of \texttt{bulk}.
These transfers also have a constant (amortized) runtime because we always transfer entire words (see below).

\paragraph{Invariants.}
We present here the simplest variant of streaming ANS, deferring generalizations to Section~\ref{sec:generalized_streaming}.
In this variant, \texttt{bulk} is a vector of \texttt{precision}-bit words (i.e., unsigned integers smaller than~$2^\texttt{precision}$), and \texttt{head} can hold up to ${2\!\times\!\texttt{precision}}$~bits (see Figure~\ref{fig:streaming-ans}).
The algorithm is very similar to our \texttt{SlowAnsCoder} from Listing~\ref{listing:slowans_inlined} except that all arithmetic operations now act on \texttt{head} instead of \texttt{compressed}, and the encoder sometimes flushes some data from \texttt{head} to \texttt{bulk} while the decoder sometimes refills data from \texttt{bulk} to \texttt{head}.
Obviously, the encoder and decoder have to agree on exactly when such data transfers occur.
This is done by upholding the following invariants:
\begin{enumerate}[(i)]
\item
    $\texttt{head} < 2^{2\times\texttt{precision}}$, i.e., \texttt{head} is a ${2\!\times\!\texttt{precision}}$-bit unsigned integer; and
\item
    $\texttt{head} \geq 2^\texttt{precision}$ if \texttt{bulk} is not empty.
\end{enumerate}
Any violation of these invariants triggers a data transfer between \texttt{head} and \texttt{bulk} that restores them.

\begin{listing}[p]
\begin{minted}[linenos,frame=lines,escapeinside=@@]{python}
class AnsCoder:
    def __init__(self, precision, compressed=[]):
        self.precision = precision
        self.mask = (1 << precision) - 1 # (a string of @$\mathtt{precision}$@ one-bits)
        self.bulk = compressed.copy() # (We will mutate @$\mathtt{bulk}$@ below.)@\label{line:ans_constructor_copy_compressed}@
        self.head = 0
        # Establish invariant (ii):
        while len(self.bulk) != 0 and (self.head >> precision) == 0:@\label{line:ans_constructor_establish_invariant1}@
            self.head = (self.head << precision) | self.bulk.pop()@\label{line:ans_constructor_establish_invariant2}@

    def push(self, symbol, m):       # Encodes one symbol.
        # Check if encoding directly onto @$\mathtt{head}$@ would violate invariant (i):
        if (self.head >> self.precision) >= m[symbol]:@\label{line:ans_check_invariant_push}@
            # Transfer one word of compressed data from @$\mathtt{head}$@ to @$\mathtt{bulk}$@:
            self.bulk.append(self.head & self.mask) # (“@{\normalfont\texttt{\&}}@” is bitwise @{\normalfont\texttt{and}}@)@\label{line:ans_transfer_push_begin}@
            self.head >>= self.precision@\label{line:ans_transfer_push_end}@
            # At this point, invariant (ii) is definitely violated,
            # but the operations below will restore it.

        z = self.head %
        self.head //= m[symbol]
        self.head = (self.head << self.precision) | z # (This is@\label{line:ans_regular_push_end}@
            # equivalent to “@{\normalfont\texttt{self.head * n + z}}@”, just slightly faster.)

    def pop(self, m):                # Decodes one symbol.
        z = self.head & self.mask    # (same as “@{\normalfont\texttt{self.head %
        self.head >>= self.precision # (same as “@{\normalfont\texttt{//= n}}@” but faster)
        for symbol, m_symbol in enumerate(m):@\label{line:ans_find_symbol_begin}@
            if z >= m_symbol:
                z -= m_symbol
            else:
                break # We found the @$\mathtt{symbol}$@ that satisfies @$\mathtt{z} \in \ZZ_i(\mathtt{symbol})$@.@\label{line:ans_find_symbol_end}@
        self.head = self.head * m_symbol + z@\label{line:ans_regular_pop_end}@

        # Restore invariant (ii) if it is violated (which happens exactly
        # if the encoder transferred data from @$\mathtt{head}$@ to @$\mathtt{bulk}$@ at this point):
        if (self.head >> self.precision) == 0 and len(self.bulk) != 0:@\label{line:ans_check_invariant_pop}@
            # Transfer data back from @$\mathtt{bulk}$@ to @$\mathtt{head}$@ (“@{\normalfont\texttt{|}}@” is bitwise @{\normalfont\texttt{or}}@):
            self.head = (self.head << self.precision) | self.bulk.pop()@\label{line:ans_transfer_pop}@

        return symbol

    def get_compressed(self):
        compressed = self.bulk.copy() # (We will mutate @$\mathtt{compressed}$@ below.)
        head = self.head
        # Chop @$\mathtt{head}$@ into @$\mathtt{precision}$@-sized words and append to @$\mathtt{compressed}$@:
        while head != 0:
            compressed.append(head & self.mask)
            head >>= self.precision
        return compressed
\end{minted}
\vspace{-4pt}
\caption{
    A complete streaming ANS entropy coder in Python.
    For a usage example, see Listing~\ref{listing:ans_usage} (replace \texttt{SlowAnsCoder} with \texttt{AnsCoder}).
    This implementation is written in Python for demonstration purpose only.
    Real deployments should be implemented in a more efficient, compiled language (see, e.g., the \texttt{constriction} library presented in Section~\ref{sec:library}, which also provides Python bindings).
}
\label{listing:ans}
\end{listing}

\paragraph{The Streaming ANS Algorithm.}
Listing~\ref{listing:ans} shows the full implementation of our final \texttt{AnsCoder}, which is an evolution on our \texttt{SlowAnsCoder} implementation from Listing~\ref{listing:slowans_inlined} (and which can be used as a replacement for \texttt{SlowAnsCoder} in the usage example of Listing~\ref{listing:ans_usage}).
Method \texttt{push} checks upfront (on Line~\ref*{line:ans_check_invariant_push}) if encoding directly onto \texttt{head} would lead to an overflow that would violate invariant~(i).
We show in Appendix~\ref{sec:appendix_correctness_conditional} that this is the case exactly if $\texttt{head} \,\texttt{>>}\, \texttt{precision} \geq m_i(x_i)$, where ``\texttt{>>}'' denotes a right bit-shift (i.e., integer division by $2^\texttt{precision})$.
If this is the case, then the encoder transfers the least significant \texttt{precision} bits from \texttt{head} to \texttt{bulk} (Lines~\ref*{line:ans_transfer_push_begin}-\ref*{line:ans_transfer_push_end}).
Since we assume that both invariants are satisfied at method entry, \texttt{head} initially contains at most ${2\!\times\!\texttt{precision}}$ bits (invariant~(i)), so transferring \texttt{precision} bits out of it leads to a \emph{temporary} violation of invariant~(ii) (but both invariants hold again on method exit as we show in Appendix~\ref{sec:appendix_correctness_invariants}).
The temporary violation of invariant~(ii) is on purpose: since transfers from \texttt{head} to \texttt{bulk} are the only operations during encoding that lead to a temporary violation of invariant~(ii), we can detect and invert such transfers on the \emph{decoder} side by simply checking for invariant~(ii), see Lines~\ref*{line:ans_check_invariant_pop}-\ref*{line:ans_transfer_pop} in Listing~\ref{listing:ans}.

The method \texttt{get\_compressed} exports the entire compressed data by concatenating \texttt{bulk} and \texttt{head} into a list of \texttt{precision}-bit words, which can be written to a file or network socket (e.g., by splitting each word into four 8-bit bytes if $\texttt{precision}=32$).
Before we analyze the \texttt{AnsCoder} in more detail, we emphasize its simplicity:
Listing~\ref{listing:ans} is a \emph{complete} implementation of a computationally efficient entropy coder with very near-optimal compression performance (see below).
By contrast, a complete implementation of arithmetic coding or range coding~\citep{rissanen1979arithmetic, pasco1976source} would be much more involved due to a number of corner cases in those algorithms.

\paragraph{Compression Performance.}
The separation between \texttt{bulk} and \texttt{head} reduces the runtime cost of ANS from quadratic to linear in the message length~$k$, but it introduces a small compression overhead.
When the encoder transfers data from \texttt{bulk} to \texttt{head} it simply chops off \texttt{precision} bits from the binary representation of \texttt{head}.
This is well-motivated as we encode onto \texttt{head} with an optimal code (with respect to~$Q$), and so one might assume that each valid bit in \texttt{head} carries one full bit of entropy.
However, this is not quite correct:
as discussed in Section~\ref{sec:fundamentals}, even with an optimal code, the expected length of the compressed bit string can exceed the entropy by up to almost $1$~bit.
Thus, each valid bit in the binary representation of \texttt{head} carries slightly less than one bit of entropy.

According to Benford's Law~\citep{hill1995statistical}, the most significant (left) bits in \texttt{head} carry lowest entropy while less significant (right) bits are nearly uniformly distributed and thus indeed carry close to one bit of entropy each.
This is why the transfer from \texttt{head} to \texttt{bulk} (Lines~\ref*{line:ans_transfer_push_begin}-\ref*{line:ans_transfer_push_end} in Listing~\ref{listing:ans}) takes the \emph{least} significant \texttt{precision} bits of \texttt{head} (see Figure~\ref{fig:streaming-ans}).
We can make their entropies arbitrarily close to one (and thus the overhead arbitrarily small) by increasing \texttt{precision} since a transfer occurs only if these bits are ``buried below'' at least an additional $\texttt{precision} - {(-\log_2 Q_i(X_i\eq x_i))}$ valid bits.

In practice, streaming ANS has very close to optimal compression performance (see empirical results in Section~\ref{sec:benchmarks} below).
But a small bitrate-overhead over a hypothetical optimal coder comes from:
\begin{enumerate}[~~~1.]
\item
    the linear (in~$k$) overhead due to Benford's Law just discussed (shrinks as \texttt{precision} increases);
\item
    a linear approximation overhead of $\sum_{i=1}^k D_\text{KL}(P_i\,||\,Q_i)$ (shrinks as \texttt{precision} increases); and
\item
    a constant overhead of at most \texttt{precision} bits due the first symbol in the bits-back trick.
\end{enumerate}
In practice, it is easy to find a \texttt{precision} that makes all three overheads negligibly small (see Section~\ref{sec:benchmarks}).
But there can be additional constraints, e.g., memory alignment, the size of lookup tables for~$Q_i$ (see Section~\ref{sec:benchmarks}), and the size of jump tables for random-access decoding (see Section~\ref{sec:random_access}).

\paragraph{Technical Details.}
Listing~\ref{listing:ans} is an educational demo implementation of streaming ANS in the Python programming language.
For efficiency, real deployments should use a language that provides more control over memory layout.
We present an efficient open-source library (with pre-compiled bindings for Python) in Section~\ref{sec:library} below.
Further, Listing~\ref{listing:ans} replaces arithmetic operations that involve $n$~$(=2^\texttt{precision})$ by equivalent but faster bitwise operations.
This allows us to avoid integer division when decoding, which is by far the slowest arithmetic operation on CPUs~\citep{fog2021instruction}.

So far, we have ignored unique decodability (see Section~\ref{sec:fundamentals}).
For a stream code like ANS, unique decodability is less important than for symbol codes since one rarely just concatenates the compressed representations of entire messages (e.g., entire compressed images) without using some container format or protocol.
But we can easily make our \texttt{AnsCoder} from Listing~\ref{listing:ans} uniquely decodable by calling its constructor with argument \texttt{compressed = [0,~1]}.
If we then encode a message, call \texttt{get\_compressed}, concatenate the result to an arbitrary existing list of words~$L'$, and finally decode the message from this concatenation, then the decoder will end up with $\texttt{bulk} = L'$ (and $\texttt{head}=n$).

This concludes our presentation of the basic algorithm for entropy coding with Asymmetric Numeral Systems~(ANS).
The next section discusses variations on this basic algorithm.
Section~\ref{sec:practice} presents and empirically evaluates an optimized open-source library of various entropy coders (including ANS).

\section{Variations on ANS}
\label{sec:beyond}

We generalize the basic streaming ANS algorithm from Section~\ref{sec:streaming-ans}.
The generalizations in Subsections~\ref{sec:generalized_streaming} and~\ref{sec:random_access} are straight-forward.
Subsection~\ref{sec:continuity} is more advanced, and it builds on our reinterpretation of ANS as bits-back coding with positional numeral systems presented in Section~\ref{sec:bits-back}.

\subsection{More General Streaming Configurations}
\label{sec:generalized_streaming}

Section~\ref{sec:streaming-ans} and Listing~\ref{listing:ans} present the simplest variant of streaming ANS.
More general variants are possible.
The bitlength of all words on \texttt{bulk} may be any integer $\texttt{word\_size} \geq \texttt{precision}$, and the \texttt{head} may have a more general bitlength $\texttt{head\_capacity} \geq \texttt{precision} + \texttt{word\_size}$.
This includes the special case from Section~\ref{sec:streaming-ans} for $\texttt{word\_size} = \texttt{precision}$ and $\texttt{head\_capacity} =2\!\times\!\texttt{precision}$, but it also admits more general setups with the following generalized invariants:
\begin{enumerate}[(i')]
    \item $\texttt{head} < 2^\texttt{head\_capacity}$, i.e., \texttt{head} can always be represented in \texttt{head\_capacity} bits; and
    \item $\texttt{head} \geq 2^{\texttt{head\_capacity} \,-\, \texttt{word\_size}}$ if \texttt{bulk} is not empty (a violation of this invariant means precisely that we can transfer one word from \texttt{bulk} to \texttt{head} without violating invariant~(i')).
\end{enumerate}

Setting \texttt{head\_capacity} larger than $\texttt{precision} + \texttt{word\_size}$ reduces overhead~1 from Section~\ref{sec:streaming-ans}.
We analyze this improvement empirically in Section~\ref{sec:benchmarks}.
Setting \texttt{word\_size} larger than \texttt{precision} may be motivated by more technical reasons, e.g., memory alignment of words on \texttt{bulk} and the memory footprint and thus cache friendliness of lookup tables for the quantile function of~$Q_i$.

\begin{listing}[t]
\begin{minted}[linenos,frame=lines,escapeinside=@@]{python}
class SeekableAnsCoder: # Adds random-access decoding to Listing @\ref*{listing:ans}@.
    # @{\normalfont\texttt{\_\_init\_\_}}@, @{\normalfont\texttt{push}}@, @{\normalfont\texttt{pop}}@, and @{\normalfont\texttt{get\_compressed}}@ same as in Listing @\ref*{listing:ans}@ ...

    def checkpoint(self): # Records a point to which we can @$\mathtt{seek}$@ later.
        return (len(self.bulk), self.head)

    def seek(self, checkpoint): # Jumps to a previously taken @$\mathtt{checkpoint}$@.
        position, head = checkpoint
        if position > len(self.bulk): # @$``\mathtt{raise}"$@ throws an exception.
            raise "This simple demo can only seek forward."
        self.bulk = self.bulk[0:position] # Truncates @$\mathtt{bulk}$@.
        self.head = head

# Usage example:
precision = 4 # For demonstration; deployments should use higher @$\mathtt{precision}$@.
m = [7, 3, 6] # Same demo model as in Listing @\ref*{listing:ans_usage}@.
coder = SeekableAnsCoder(precision)
message = [2, 0, 2, 1, 0, 1, 2, 2, 2, 1, 0, 2, 1, 2, 0, 0, 1, 1, 1, 2]

for symbol in reversed(message[10:20]): # Encode second half of @$\mathtt{message}$@.
    coder.push(symbol, m)
checkpoint = coder.checkpoint()         # Record a checkpoint.
for symbol in reversed(message[0:10]):  # Encode first half of @$\mathtt{message}$@.
    coder.push(symbol, m)

assert coder.pop(m) == message[0]       # Decode first symbol.
assert coder.pop(m) == message[1]       # Decode second symbol.
coder.seek(checkpoint)                  # Jump to 11th symbol.
assert [coder.pop(m) for _ in range(10)] == message[10:20] # Decode rest.
\end{minted}
\vspace{-4pt}
\caption{
    Streaming ANS with random-access decoding (see Section~\ref{sec:random_access}).
    This simple demo implementation can only seek forward since both seeking and decoding \emph{consume} compressed data.
    To allow seeking backwards, one could use a cursor into immutable compressed data instead.
}
\label{listing:random_access}
\end{listing}

\subsection{Random-Access Decoding}
\label{sec:random_access}

When decoding a compressed message, it is sometimes desirable to quickly jump to some specific position within the message (e.g., when skipping forward in a compressed video stream).
Such random access is trivial in symbol codes as the decoder only needs to know the correct offset into the compressed bit string, which may be stored for selected potential jump targets as meta data (``jump table'') within some container format.
For a stream code like ANS, jumping (or ``seeking'') to a specific position within the message requires knowledge not only of the offset within the compressed bit string but also of the internal state that the decoder would have if it arrived at this position without seeking.
ANS makes it particularly easy to obtain this target decoder state during encoding since---as opposed to, e.g., arithmetic coding---the encoder and decoder in ANS are the same data structure with the same internal state (i.e.,~\texttt{head}).
Listing~\ref{listing:random_access} shows an example of random access with ANS.

\subsection{Continuity With Respect to Entropy Models}
\label{sec:continuity}

We now discuss a more specialized variation on ANS that may not be immediately relevant to most readers, but which demonstrates how useful a deep understanding of the ANS algorithm can be for developing new ideas.
This is an advanced chapter.
First time readers might prefer to skip to Section~\ref{sec:practice} for empirical results and practical advice in more common use cases before coming back here.

Streaming ANS as presented in Listing~\ref{listing:ans} is a highly effective algorithm for entropy coding with a fixed model.
But the situation becomes tricker if modeling and entropy coding cannot be separated as clearly.
For example, novel deep-learning based compression methods often employ probabilistic models over \emph{continuous} spaces, and thus the models have to be discretized in some way before they can be used as entropy models.
Advanced discretization methods might benefit from an end-to-end optimization on the encoder side that optimizes through the entropy coder.
Unfortunately, optimizing a function that involves an entropy coder like ANS is extremely difficult since ANS packs as much information content into as few bits as possible, which leads to highly noncontinuous behavior.

\begin{listing}[t]
\begin{minted}[linenos,frame=lines,escapeinside=@@]{python}
precision = 4 # For demonstration; deployments should use higher @$\mathtt{precision}$@.
m_orig = [7, 3, 6] # Same demo entropy model as in Listing @\ref*{listing:ans_usage}@.
m_mod  = [6, 4, 6] # (Slightly) modified entropy model compared to @{\normalfont\texttt{m\_orig}}@.
compressed = [0b1001, 0b1110, 0b0110, 0b1110] # Some example bit string.

# Case 1: decode 4 symbols using entropy model @{\normalfont\texttt{m\_orig}}@ for all symbols:
decoder = AnsCoder(precision, compressed) # @$\mathtt{AnsCoder}$@ defined in Listing @\ref*{listing:ans}@.
case1 = [decoder.pop(m_orig) for _ in range(4)]

# Case 2: change the entropy model, but *only* for the first symbol:
decoder = AnsCoder(precision, compressed) # “@$\mathtt{compressed}$@” hasn't changed.
case2 = [decoder.pop(m_mod)] + [decoder.pop(m_orig) for _ in range(3)]

print(f"case1 = {case1}") # Prints: “case1 = [0, 1, 0, 2]”
print(f"case2 = {case2}") # Prints: “case2 = [1, 1, 2, 0]”
                          #                   @$\;\!\uparrow$@     @$\;\nnwarrow$@  @$\nnwarrow$@
                          #   We changed only the  @$\mathtt{|}$@ But that affected both
                          # model for this symbol. @$\mathtt{|}$@ of these symbols too.
\end{minted}
\vspace{-4pt}
\caption{
    Non-local effects of entropy models (see Section~\ref{sec:continuity});
    changing the entropy model for the first decoded symbol of a sequence leads to a ripple effect that may affect all subsequent symbols.
}
\label{listing:chain_coder_motivation}
\end{listing}

For example, Listing~\ref{listing:chain_coder_motivation} demonstrates that ANS reacts in a very irregular and non-local way to even tiny changes of the entropy model.
The example decodes a sequences of symbols from an example bit string \texttt{compressed}.
Such a sequence of decoding operations could be part of a higher-level \emph{encoder} that employs the bits-back trick on some latent variable model with a high-dimensional latent space.
In this case, the encoder would still be allowed to optimize over certain model parameters (or, e.g., discretization settings) provided that it appends the final choice of those parameters to the compressed bit string before transmitting it.
To demonstrate the effect of optimizing over model parameters, Listing~\ref{listing:chain_coder_motivation} decodes from the same bit string twice.%
\footnote{Although lists are passed by reference in Python, the first decoder in Listing~\ref{listing:chain_coder_motivation} does not mutate \texttt{compressed} since the constructor of \texttt{AnsCoder} performs a \texttt{copy} of the provided bit string, see Line~\ref*{line:ans_constructor_copy_compressed} in Listing~\ref{listing:ans}.}
The employed entropy models differ slightly between these two iterations, but \emph{only for the first symbol}.
Yet, the decoded sequences differ not only on the first symbol but also on subsequent symbols that were decoded with identical entropy models.

This ripple effect is no surprise:
changing the entropy model for the first symbol affects not only the immediately decoded symbol but also the internal coder state after decoding the first symbol, which in turn affects how the coder proceeds to decoding subsequent symbols.
Our deeper understanding of ANS as a form of bits-back coding (see Section~\ref{sec:bits-back}) allows us to pinpoint the problem more precisely as it splits the process of decoding a symbol~$x_i$ into three steps:
(1)~decoding a number~$z_i$ from the compressed data using the (fixed) alphabet $\{0,\ldots,{n-1}\}$;
(2)~identifying the unique symbol $x_i$ that satisfies $z_i\in\ZZ_i(x_i)$; and
(3)~encoding~$z_i$ back onto the compressed data, this time using the (model-dependent) alphabet~$\ZZ_i(x_i)$.
Note that step~(1) is independent of the entropy model.
The only reason why changing the entropy model for one symbol affects subsequent symbols is step~(3), which ``leaks'' information about the current entropy model to the stack of compressed data.

\begin{listing}[p]
\begin{minted}[linenos,frame=lines,escapeinside=@@]{python}
class ChainCoder: # Prevents the non-local effect shown in Listing @\ref*{listing:chain_coder_motivation}@.
    def __init__(self, precision, compressed, remainders=[]):
        """Initializes a ChainCoder for decoding from `compressed`."""
        self.precision = precision
        self.mask = (1 << precision) - 1
        self.compressed = compressed.copy() # @$\mathtt{pop}$@ decodes from here.
        self.remainders = remainders.copy() # @$\mathtt{pop}$@ encodes onto here.
        self.remainders_head = 0
        # Establish invariant (ii):
        while len(self.remainders) != 0 and \
                (self.remainders_head >> precision) == 0:
            self.remainders_head <<= self.precision
            self.remainders_head |= self.remainders.pop()

    def pop(self, m): # Decodes one symbol.
        z = self.compressed.pop() # Always read a full word from @$\mathtt{compressed}$@.@\label{line:chain_pop_step1}@
        for symbol, m_symbol in enumerate(m):@\label{line:chain_pop_step2_begin}@
            if z >= m_symbol:
                z -= m_symbol
            else:
                break # We found the @$\mathtt{symbol}$@ that satisfies @$\mathtt{z} \in \ZZ_i(\mathtt{symbol})$@.@\label{line:chain_pop_step2_end}@

        self.remainders_head = self.remainders_head * m_symbol + z@\label{line:chain_pop_step3_begin}@
        if (self.remainders_head >> (2 * self.precision)) != 0:
            # Invariant (i) is violated. Flush one word to @{\normalfont\texttt{remainders}}@.
            self.remainders.append(self.remainders_head & self.mask)
            self.remainders_head >>= self.precision
            # It can easily be shown that invariant (i) is restored here.@\label{line:chain_pop_step3_end}@

        return symbol

    def push(self, symbol, m): # Encodes one symbol.
        if len(self.remainders) != 0 and \
                self.remainders_head < (m[symbol] << self.precision):
            self.remainders_head <<= self.precision
            self.remainders_head |= self.remainders.pop()
            # Invariant (i) is now violated but will be restored below.

        z = self.remainders_head %
        self.remainders_head //= m[symbol]
        self.compressed.append(z)
\end{minted}
\vspace{-4pt}
\caption{
    Sketch of an entropy coder that is similar to ANS but that prevents the non-local effect demonstrated in Listing~\ref{listing:chain_coder_motivation} by using separate stacks of data for reading and writing (see Section~\ref{sec:continuity}).
}
\label{listing:chain_coder}
\end{listing}

This detailed understanding allows us to come up with an alternative entropy coder that does not exhibit the ripple effect shown in Listing~\ref{listing:chain_coder_motivation}.
We can prevent the coder from leaking information about entropy models from one symbol to the next by using two separate stacks for the decoding and encoding operations in steps (1) and~(3) above.
Listing~\ref{listing:chain_coder} sketches an implementation of such an entropy coder.
The method \texttt{pop} decodes $z_i\in\{0,\ldots,{n-1}\}$ from \texttt{compressed} (Line~\ref*{line:chain_pop_step1}), identifies the symbol~$x_i$ (Lines~\ref*{line:chain_pop_step2_begin}-\ref*{line:chain_pop_step2_end}), and then encodes $z_i\in\ZZ_i(x_i)$ onto a \emph{different} stack \texttt{remainders} (Lines~\ref*{line:chain_pop_step3_begin}-\ref*{line:chain_pop_step3_end}).
If we use this coder to decode a sequence of symbols (as in the example of Listing~\ref{listing:chain_coder_motivation}), then any changes to an entropy model for a single symbol affect only that symbol and the data that accumulates on \texttt{remainders}, but it has no effect on any subsequently decoded symbols.
Once all symbols are decoded, one can concatenate \texttt{compressed} and \texttt{remainders} in an appropriate way.

As a technical remark, Listing~\ref{listing:chain_coder} implements the simplest streaming configuration for such an entropy coder, with $\texttt{word\_size} = \texttt{precision}$ and $\texttt{head\_capacity} = 2\!\times\!\texttt{precision}$.
More general configurations analogous to Section~\ref{sec:generalized_streaming} are possible and left as an exercise to the reader (hint: setting $\texttt{word\_size} > \texttt{precision}$ requires introducing a separate \texttt{compressed\_head}).

In summary, our new perspective on ANS as bits-back coding with a \texttt{UniformCoder} allowed us to resolve a non-locality issue of ANS.
While this specific issue is unlikely to be immediately relevant to most readers, other limitations of ANS might, and being able to semantically split ANS into the three subtasks of the bits-back trick might help overcome those too.
This concludes our discussion of variations and generalizations of ANS.
The next section provides some practical guidance on the streaming configuration based on empirically observed compression performances and runtime costs.

\section{Software Library and Empirical Results}
\label{sec:practice}

In this section, we briefly present an open-source library of various entropy coders (including ANS), we compare empirical bitrates and runtime costs for compressing real-world data across entropy coders and their configurations, and we provide practical advice based on these empirical results.

\subsection{A Library of Entropy Coders for Research and Production}
\label{sec:library}

Data compression combines the procedural (i.e., algorithmic) task of entropy coding with the declarative task of probabilistic modeling.
Recently, this separation of tasks has started to manifest itself also in the division of labor among researchers:
systems and signal processing researchers are continuing to optimize compression algorithms for real hardware while machine learning researchers have started to introduce new ideas for the design of powerful probabilistic models~(e.g., \citep{toderici2017full,minnen2018joint,agustsson2020scale,yang2020hierarchical,yang2021insights}).
Unfortunately, these two communities traditionally use vastly different software stacks, which seems to be slowing down idea transfer and thus might be part of the reason why machine-learning based compression methods are still hardly used in the wild despite their proven superior compression performance.

Along with this paper, the author is releasing \texttt{constriction},%
\footnote{\url{https://bamler-lab.github.io/constriction}}
an open-source library of entropy coders that intends to bridge the gap between systems and machine learning researchers by providing first-class support for both the Python and Rust programming language.
To demonstrate \texttt{constriction} in a real deployment, we point readers to The Linguistic Flux Capacitor,%
\footnote{\url{https://robamler.github.io/linguistic-flux-capacitor}\label{footnote:tlfc}}
a web application that uses \texttt{constriction} in a WebAssembly module to execute the neural compression method proposed in~\citep{yang2020variational} completely on the client side, i.e., in the user's web browser.

Machine learning researchers can use \texttt{constriction} in their typical workflow through its Python API.
It provides a consistent interface to various entropy coders, but it deliberately dictates the precise configuration of each coder in a way that prioritizes compression performance (at the cost of some runtime efficiency, see Section~\ref{sec:benchmarks}).
This allows machine learning researchers to quickly pick a coder that fits their model architecture (e.g, a range coder for autoregressive models, or ANS for bits-back coding with latent variable models) without having to learn a new way of representing entropy models and bit strings for each coder type, and without having to research complex coder configurations.

Once a successful prototype of a new compression method is implemented and evaluated in Python, \texttt{constriction}'s Rust API simplifies the process of turning it into a self-contained product (i.e., an executable, static library or WebAssembly module).
By default, \texttt{constriction}'s Rust and Python APIs are binary compatible, i.e., data encoded with one can be decoded with the other, which simplifies debugging.
In addition, the Rust API optionally admits control (via compile-time type parameters) over coder details (streaming configurations, see Section~\ref{sec:generalized_streaming}, and data providers).
This allows practitioners to tune a working prototype for optimal runtime and memory efficiency.
The next section provides some guidance for these choices by comparing empirical bitrates and runtime costs.

\subsection{Empirical Results and Practical Advice}
\label{sec:benchmarks}

We analyze the bitrates and runtime cost of Asymmetric Numeral Systems (ANS) with various streaming configurations (see Section~\ref{sec:generalized_streaming}), and we compare to Arithmetic Coding~(AC)~\citep{rissanen1979arithmetic, pasco1976source} and Range Coding (RC)~\citep{martin1979range}.
We identify a streaming configuration where both ANS and RC are consistently effective over a large range of entropies, and we observe that that they are both considerably faster than AC.
Based on these findings, we provide practical advice for when and how to use which entropy coding algorithm.

\paragraph{Benchmark Data.}
We test all entropy coders on the data that are decoded in The Linguistic Flux Capacitor web app (see footnote~\ref*{footnote:tlfc} on page \pageref{footnote:tlfc}), which are the discretized model parameters of the natural language model from~\citep{bamler2017dynamic}.
We chose these discretized model parameters as benchmark data because they arose in a real use case and they cover a wide spectrum of entropies per symbol spanning four orders of magnitude.
The model parameters consist of 209~slices of 3~million parameters each.
We treat the symbols within each slice as i.i.d.~and use the empirical symbol frequencies within each slice as the entropy model when encoding and decoding the respective slice.
Before discretization, the model parameters were transformed such that each slice represents the difference from other slices of varying distance (similar to I- and B-frames in a video codec).
This leads to a wide range of entropies, ranging from about $0.001$ to~$10$~bits per symbol from the lowest to highest entropy slice (with an overall average of $0.41$~bits per symbol).

\paragraph{Experiment Setup.}
We use the proposed \texttt{constriction} library through its Rust API for both ANS and RC, and the third-party Rust library \texttt{arcode}~\citep{burgess2021arcode} for AC.
In the following, we specify the streaming configuration (see Section~\ref{sec:generalized_streaming}) for ANS and RC by the triple ``$\texttt{precision}\,/\,\texttt{word\_size}\,/\,\texttt{head\_capacity}$''.
These parameters can be set in \texttt{constriction}'s Rust API at compile time so that the compiler can generate optimized code for each configuration.
The library defines two presets: ``default'' (${24\,/\,32\,/\,64}$), which is recommended for initial prototyping (and therefore exposed through the Python API), and ``small'' (${12\,/\,16\,/\,32}$), which is optimized for runtime and memory efficiency.
In addition to these two preset configurations, we also experiment with ${32\,/\,32\,/\,64}$ and ${16\,/\,16\,/\,32}$, which  match the simpler setup of Section~\ref{sec:streaming-ans} (i.e., $\texttt{precision}=\texttt{word\_size}$).
For AC, the only tuning parameter is the fixed-point \texttt{precision}.
We report results with $\texttt{precision}=63$ (the highest precision supported by \texttt{arcode}) since lowering the precision did not improve runtime performance.
Our benchmark code and data is available online.%
\footnote{\url{https://github.com/bamler-lab/understanding-ans}}

The reported runtime for each slice is the median over 10 measurements on an Intel Core i7-7500U CPU (2.70~GHz) running Ubuntu Linux~21.10.
We ran an entire batch of experiments (encoding and decoding all slices with all tested entropy coders) in the inner loop and repeated this procedure 10~times, randomly shuffling the order of experiments for each of the 10~batches.
For each experiment, we encoded/decoded the entire slice twice in a row and measured only the runtime of the second run, so as to minimize variations in memory caches and branch predictions.
We calculated and saved a trivial checksum (bitwise \texttt{xor}) of all decoded symbols to ensure that no parts of the decoding operations were optimized out.
The benchmark code was compiled with \texttt{constriction} version~0.1.4, \texttt{arcode} version~0.2.3, and Rust version~1.57 in \texttt{-{}-release} mode and ran in a single thread.

\begin{table}
    \caption{
        Empirical compression performances and runtimes of Asymmetric Numeral Systems~(ANS), Range Coding~(RC), and Arithmetic Coding~(AC);
        ANS and RC use the proposed \texttt{constriction} library, which admits arbitrary streaming configurations (see Section~\ref{sec:generalized_streaming}).
        Python bindings are provided for both ANS and RC for the configurations labeled as ``default'' in this table, which both have negligible~($<0.1\,\%$) bitrate overhead while being considerably faster than arithmetic coding.
    }
    \label{table:results_summary}
    \centering
    \begin{tabularx}{\textwidth}{lcclYY}
        \toprule
        Method $(\texttt{precision} \,/$ & \multicolumn{2}{c}{Bitrate overhead over Eq.~\ref{eq:source-coding-theorem-lower}} && \multicolumn{2}{c}{Runtime [ns per symbol]} \\
        $\texttt{word\_size} \,/\, \texttt{head\_capacity})$ & $\blacktriangleright$ total & $\blacktriangleright\!D_\text{KL}(P \,||\, Q)$ && $\blacktriangleright$ encoding & $\blacktriangleright$ decoding \\
        \midrule
        ANS $(24\,/\,32\,/\,64)$ \emph{(``default'')}$\quad$
            & $0.0015\,\%$ & $2.6\!\times\! 10^{-6}\,\%$ && $24.2$ & $~~6.1$ \\
        ANS $(32\,/\,32\,/\,64)$
            & $0.0593\,\%$ & $<10^{-8}\,\%$         && $24.2$ & $~~6.9$ \\
        ANS $(16\,/\,16\,/\,32)$
            & $0.2402\,\%$ & $0.1235\,\%$              && $19.8$ & $~~6.4$ \\
        ANS $(12\,/\,16\,/\,32)$ \emph{(``small'')}
            & $3.9567\,\%$ & $3.9561\,\%$              && $19.8$ & $~~6.9$ \\
        \midrule
        RC $(24\,/\,32\,/\,64)$ \emph{(``default'')}
            & $0.0237\,\%$ & $2.6\!\times\! 10^{-6}\,\%$ && $16.6$ & $14.3$ \\
        RC $(32\,/\,32\,/\,64)$
            & $1.6089\,\%$ & $<10^{-8}\,\%$         && $16.7$ & $14.8$ \\
        RC $(16\,/\,16\,/\,32)$
            & $3.4950\,\%$ & $0.1235\,\%$              && $16.9$ & $~~9.4$ \\
        RC $(12\,/\,16\,/\,32)$ \emph{(``small'')}
            & $4.5807\,\%$ & $3.9561\,\%$              && $16.8$ & $~~9.4$ \\
        \midrule
        Arithmetic Coding (AC)
            & $0.0004\,\%$ & n/a                     && $43.2$ & $85.6$ \\
        \bottomrule
    \end{tabularx}
\end{table}

\begin{figure}%
    \centering%
    \includegraphics[width=\textwidth]{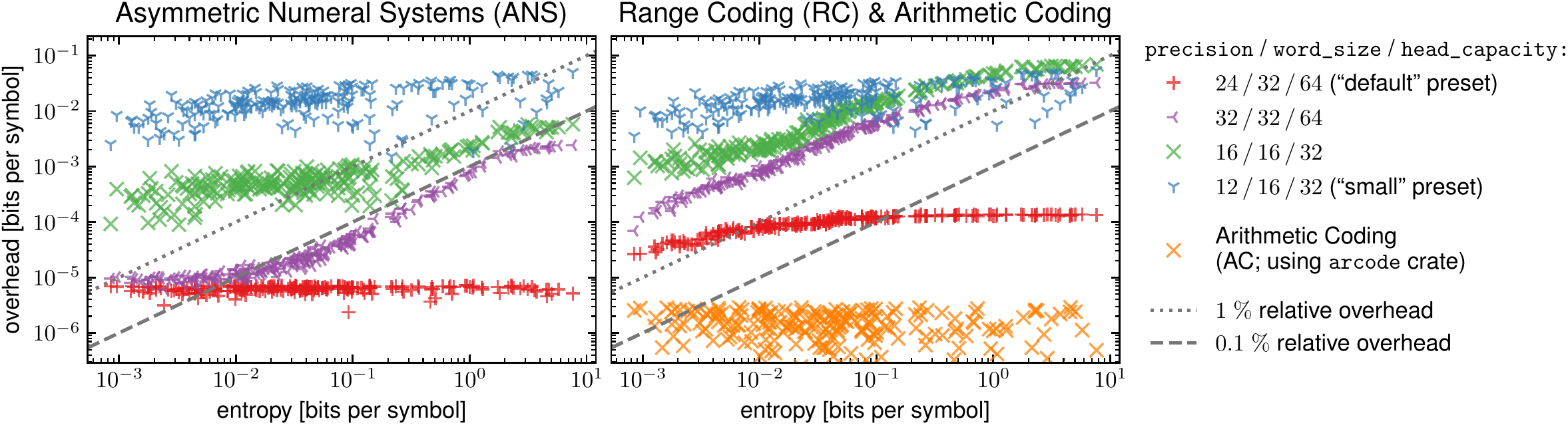}%
    \caption{%
        Empirical compression performances (bitrate overhead over Eq.~\ref{eq:source-coding-theorem-lower}) of various entropy coders as a function of entropy per symbol in each slice of the benchmark data.
        The ``default'' presets (red) for ANS and RC have a consistently low overhead over a wide range of entropies.
        The ``small'' presets (blue) may be advantageous in memory or runtime-constrained situations (see also Figure~\ref{fig:runtimes}).
    }%
    \label{fig:bitrates}%
\end{figure}

\paragraph{Results 1: Bitrates.}
Table~\ref{table:results_summary} shows aggregate empirical results over all slices of the benchmark data;
A bitrate overhead of $1\,\%$ in the second column would mean that the number of bits produced by a coder when encoding the entire benchmark data is $1.01$~times the total information content of the benchmark data.
We observe that arithmetic coding has the lowest bitrate overhead as expected, but the ``default'' preset for both ANS and RC in \texttt{constriction} has negligible~($<0.1\,\%$) overhead too on this data set.
Interestingly, setting $\texttt{precision}$ slightly smaller than $\texttt{word\_size}$ is beneficial for overall compression performance, suggesting a Benford's Law contribution to the overhead (see Section~\ref{sec:streaming-ans}).
Reducing the \texttt{precision} increases the overhead, and the third column of Table~\ref{table:results_summary} reveals that the overhead in the low-\texttt{precision} regime is largely due to the approximation error $D_\text{KL}(P\,||\,Q)$.
Figure~\ref{fig:bitrates} breaks down the overhead for each of the 209~slices in the data set, plotting them as a function of the entropy in each slice.
We observe that the ``default'' preset (red plus markers) provides the consistently best bitrates within each method over a wide range of entropies, which is why \texttt{constriction}'s Rust and Python APIs guide users to use this preset for initial prototyping.

\begin{figure}%
    \centering%
    \includegraphics[width=\textwidth]{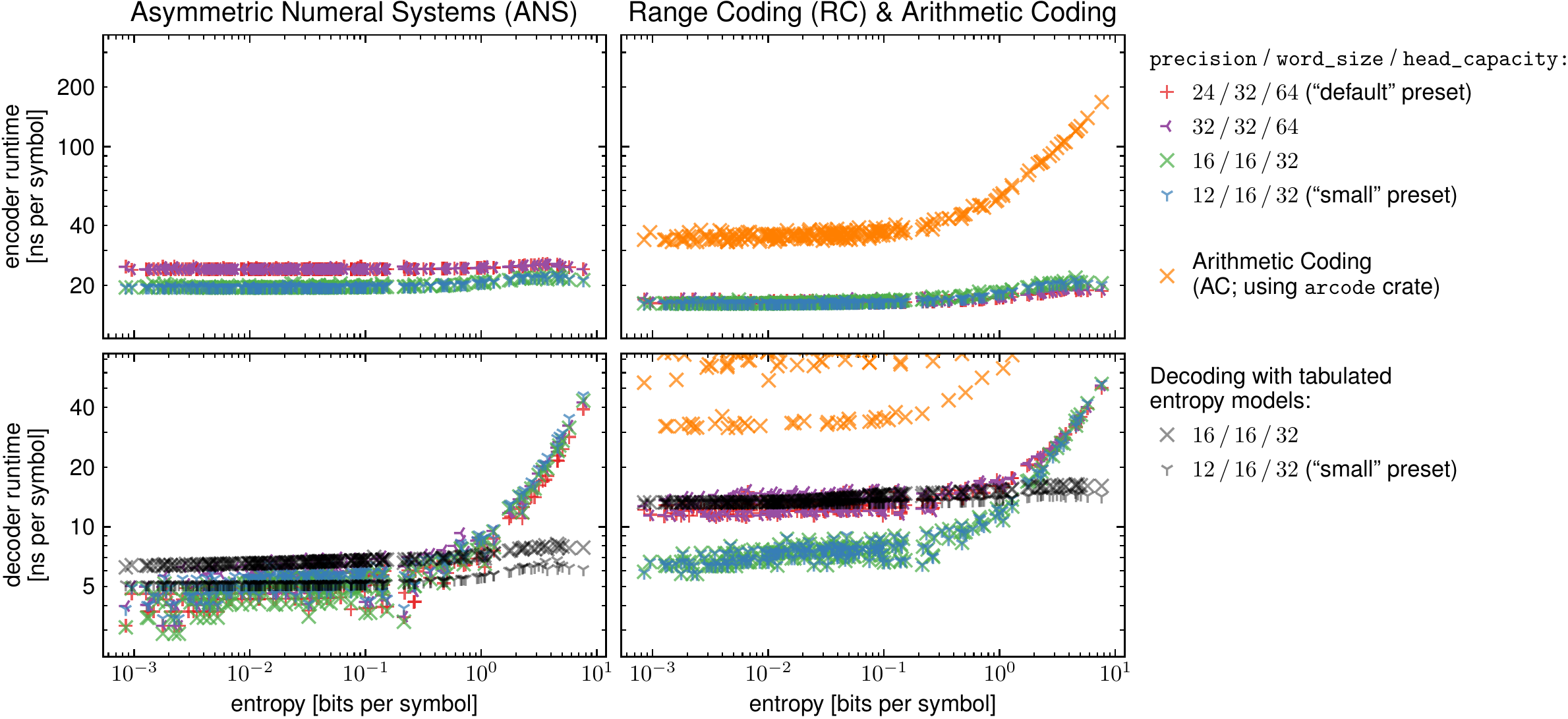}%
    \caption{%
        Measured runtimes for encoding (top) and decoding (bottom) as a function of the entropy per symbol.
        Models with low \texttt{precision} admit tabulating the mapping $z_i \mapsto x_i: z_i\in\ZZ_i(x_i)$ for all $z_i\in\{0, \ldots, {2^\texttt{precision}-1}\}$, which speeds up decoding in the high-entropy regime (gray markers).
    }%
    \label{fig:runtimes}%
\end{figure}

\paragraph{Results 2: Runtimes.}
The last two columns in Table~\ref{table:results_summary} list the runtime cost (in nanoseconds per symbol) for encoding and decoding, averaged over the entire benchmark data.
We observe that ANS provides the fastest decoder while RC provides the fastest encoder.
In the ``default'' preset, decoding with ANS is more than twice as fast compared to RC.
AC is much slower than both ANS and RC for both encoding and decoding in our experiments.
While the precise runtimes reported here should be taken with a grain of salt since they depend on the specific implementation of each algorithm, one can generally expect AC to be considerably slower than ANS and RC since it reads and writes compressed data bit by bit, which is a poor fit for modern hardware.
Indeed, Figure~\ref{fig:runtimes}, which plots runtimes as a function of entropy in the respective slice of the benchmark data, reveals that AC (orange crosses) slows down in the high-entropy regime, i.e., when it has to read or write many bits per symbol.

While the encoder runtimes for ANS and RC (upper panels in Figure~\ref{fig:bitrates}) depend only weakly on the entropy, decoding with ANS and RC slows down in the high-entropy regime.
The main computational cost in this regime turns out to be the task of identifying the symbol~$x_i$ that satisfies $z_i\in\ZZ_i(x_i)$ (corresponding to Lines~\ref*{line:ans_find_symbol_begin}-\ref*{line:ans_find_symbol_end} in Listing~\ref{listing:ans}).
For categorical entropy models like the ones used here, \texttt{constriction} uses a binary search here by default, which tends to require more iterations in models with high entropy.
The library therefore provides an alternative way to represent categorical entropy models that tabulates the mapping $z_i\mapsto x_i$ for all $z_i\in\{0,\ldots,n-1\}$ upfront, leading to a constant lookup time.
Since the size of such lookup tables is proportional to $n=2^\texttt{precision}$, they are only viable for low \texttt{precision}s (gray crosses and Y-shaped markers in the bottom panels in Figure~\ref{fig:runtimes}).
We observe that tabulated entropy models indeed speed up decoding in the high-entropy regime but they come at a cost in the low entropy regime (likely because large lookup tables are not cache friendly).

\paragraph{Practical Advice.}
In summary, the Asymmetric Numeral Systems (ANS) algorithm reviewed in this paper, as well as Range Coding (RC), both provide very close to optimal bitrates while being considerably faster than Arithmetic Coding.
While the precise bitrates and runtimes differ slightly between ANS and RC, in practice, it is usually more important to pick an entropy coder that fits the model architecture: RC operates as a queue (first-in-first-out), which simplifies compression with autoregressive models, while ANS operates as a stack (last-in-first-out), which simplifies bits-back coding with latent variable models.
Both ANS and RC can be configured by parameters that trade off compression performance against runtime and memory efficiency.
Users of the \texttt{constriction} library are advised to use the configuration from the ``default'' preset for initial prototyping (as guided by the API), and to tune the configuration only once they have implemented a working prototype.

\section{Conclusion}
\label{sec:conclusion}

We provided an educational discussion of the Asymmetric Numeral Systems (ANS) entropy coder, explaining the algorithm's internals using concepts that are familiar to many statisticians and machine-learning scientists.
This allowed us to understand Asymmetric Numeral Systems as a generalization of Positional Numeral Systems, and as an application of the bits-back trick.
Splitting up ANS into the three distinct steps of the bits-back trick allowed us to generalize the method by combining these three steps in new ways.
We hope that more idea transfer like this between the procedural (algorithmic) and the declarative (modeling) subcommunities within the field of compression research will spark novel ideas for compression methods in the future.

From a more practical perspective, we presented \texttt{constriction}, a new open-source software library that provides a collection of effective and efficient entropy coders and adapters for defining complex entropy models.
The library is intended to simplify compression research (by providing a catalog of several different entropy coders within a single consistent framework) and to speed up the transition from research code to self-contained software products (by providing binary compatible entropy coders for both Python and Rust).
We showed empirically that the entropy coders in \texttt{constriction} have very close to optimal compression performance while being much faster than Arithmetic Coding.

\section*{Acknowledgments}

The author thanks Tim Zhenzhong Xiao for stimulating discussions and Stephan Mandt for important feedback.
This work was supported by the German Federal Ministry of Education and Research (BMBF): Tübingen AI Center, FKZ: 01IS18039A.
Robert Bamler is a member of the Machine Learning Cluster of Excellence, funded by the Deutsche Forschungsgemeinschaft (DFG, German Research Foundation) under Germany’s Excellence Strategy – EXC number 2064/1 – Project number 390727645.
The author thanks the International Max Planck Research School for Intelligent Systems (IMPRS-IS) for support.

\bibliography{references}

\newpage
\appendix
\setcounter{page}{1}
\setcounter{equation}{0}
\renewcommand{\thepage}{A\arabic{page}}
\renewcommand{\theequation}{A\arabic{equation}}
\renewcommand{\thetable}{A\arabic{table}}
\renewcommand{\thefigure}{A\arabic{figure}}

\section{Appendix: Proof of Correctness of Streaming ANS}
\label{sec:appendix_correctness}

We prove that the \texttt{AnsCoder} from Listing~\ref{listing:ans} of the main text implements a correct encoder/decoder pair, i.e., that its methods \texttt{push} and \texttt{pop} are inverse to each other.
The proof uses that the \texttt{AnsCoder} upholds the two invariants (i) and~(ii) from Section~\ref{sec:streaming-ans} of the main text, which we also prove.

\paragraph{Assumptions and Problem Statement.}
We assume that the \texttt{AnsCoder} is used correctly:
the constructor (``\texttt{\_\_init\_\_}'') must be called with a positive integer \texttt{precision} and its optional argument \texttt{compressed}, if given, must be a list of nonnegative integers%
\footnote{As a side remark, note that any trailing zero words on \texttt{compressed} are meaningless and will not be returned by \texttt{get\_compressed}.
Thus, if the compressed data comes from an external source that may produce data with trailing zero words then one should always append a single ``$1$''~word to the compressed data before passing it to the constructor of \texttt{AnsCoder} so that one can reliably recover the original data, including any trailing zero words, from the return value of \texttt{get\_compressed}.
The Rust API of the \texttt{constriction} library proposed in Section~\ref{sec:library} provides the alternative constructor \texttt{from\_binary} for precisely this use case.}
that are all smaller than $n=2^\texttt{precision}$).
Once the \texttt{AnsCoder} is constructed, its methods \texttt{push} and \texttt{pop} may be called in arbitrary order and repetition to bring the \texttt{AnsCoder} into some original state~$S_0$.
In this initial setup phase that establishes state~$S_0$, the employed entropy models (argument~\texttt{m} in both \texttt{push} and \texttt{pop}) may vary arbitrarily across method calls and do not have to match between any of the \texttt{push} and \texttt{pop} calls, but they do have to specify valid entropy models (i.e., each \texttt{m} must be a nonempty list of nonnegative integers that sum to~$n$).
Further, the argument \texttt{symbol} of each \texttt{push} call must be a nonnegative integer that is smaller than the length of the corresponding~\texttt{m}, and it must satisfy $\texttt{m[symbol]} \neq 0$ (ANS cannot encode symbols with zero probability under the approximated entropy model~$Q_i$).
Once state~$S_0$ is established, we consider two scenarios:
\begin{enumerate}[(a)]
\item we call \texttt{push(symbol, m)} with an arbitrary $\texttt{symbol} \in \{0,\ldots, \texttt{len(m)}\}$, followed by \texttt{pop(m)} with the same (valid) model~\texttt{m}; or \label{item:appendix_correctness_scenario_pushpop}
\item we call \texttt{pop(m)}, assign the return value to a variable \texttt{symbol}, and then call \texttt{push(symbol, m)} with the returned \texttt{symbol} and the same (valid) model~\texttt{m}. \label{item:appendix_correctness_scenario_poppush}
\end{enumerate}
We show that, after executing either of the above two scenarios, the \texttt{AnsCoder} ends up back in state~$S_0$.
For scenario~(\ref*{item:appendix_correctness_scenario_pushpop}), we further show that the final \texttt{pop} returns the \texttt{symbol} that was used in the preceding \texttt{push}.
In Subsection~\ref{sec:appendix_correctness_conditional}, we prove these claims under the assumption that the two invariants (i) and~(ii) from Section~\ref{sec:streaming-ans} of the main text hold at entry of all methods.
In Subsection~\ref{sec:appendix_correctness_invariants}, we show that \texttt{AnsCoder} indeed upholds these two invariants (for completeness, the invariants are: (i)~$\texttt{head} < n^2$ (always) and (ii)~$\texttt{head} \geq n$ if \texttt{bulk} is not empty, where $n=2^\texttt{precision}$).

\subsection{Proof of Correctness, Assuming Invariants are Upheld}
\label{sec:appendix_correctness_conditional}

To simplify the discussion, we conceptually split the method \texttt{push} into two parts, \texttt{conditional\_flush} (Lines~\ref*{line:ans_check_invariant_push}-\ref*{line:ans_transfer_push_end} in Listing~\ref{listing:ans} of the main text) followed by \texttt{push\_onto\_head} (Lines~\ref*{line:ans_regular_push_begin}-\ref*{line:ans_regular_push_end}).
Similarly, we conceptually split the method \texttt{pop} into \texttt{pop\_from\_head} \hbox{(Lines~\ref*{line:ans_regular_pop_begin}-\ref*{line:ans_regular_pop_end})} followed by \texttt{conditional\_refill} \hbox{(Lines~\ref*{line:ans_check_invariant_pop}-\ref*{line:ans_transfer_pop})}.
The parts \texttt{push\_onto\_head} and \texttt{pop\_from\_head} are analogous to the \texttt{push} and \texttt{pop} method, respectively, of the \texttt{SlowAnsCoder} (Listing~\ref{listing:slowans_inlined}), and it is easy to see that they are inverse to each other (regardless of whether or not invariants (i) and~(ii) hold).
The tricker part is to show that the conditional flushing and refilling happen consistently, i.e., either none or both of them occur.

\paragraph{Scenario~(\ref*{item:appendix_correctness_scenario_pushpop}):}
Scenario~(\ref*{item:appendix_correctness_scenario_pushpop}) executes the sequence \texttt{conditional\_flush}, \texttt{push\_onto\_head}, \texttt{pop\_from\_head}, \texttt{conditional\_refill}.
Since the second and third step are inverse to each other, this simplifies to:
starting from the original state~$S_0$;
then executing \texttt{conditional\_flush}, which brings the \texttt{AnsCoder} into a new state~$S'$;
and then executing \texttt{conditional\_refill}.
We show that the final state is again~$S_0$ (this also shows that the \texttt{symbol} returned by the final \texttt{pop} is the one used in the call to \texttt{push}, since the problem effectively reduces to a \texttt{SlowAnsCoder}).
Both \texttt{conditional\_flush} and \texttt{conditional\_refill} are predicated on some condition (see 
\texttt{if}-statements on Lines~\ref*{line:ans_check_invariant_push} and~\ref*{line:ans_check_invariant_pop}, respectively), and each part is a no-op if its respective condition is not met.
Thus, if the condition for \texttt{conditional\_flush} is not satisfied, then $S'=S_0$, which satisfies invariant~(ii) by assumption.
The condition for \texttt{conditional\_refill} checks precisely for violation of invariant~(ii), so \texttt{conditional\_refill} also becomes a no-op in this case, and we trivially end up in state~$S_0$.
If, on the other hand, the condition for \texttt{conditional\_flush} is satisfied, then it is easy to see that $S'$ violates invariant~(ii):
line~\ref*{line:ans_transfer_push_begin} ensures that \texttt{bulk} is not empty and line~\ref*{line:ans_transfer_push_end} performs a right bit-shift of \texttt{head} by \texttt{precision}, which sets $\texttt{head}_{S'} = \lfloor \texttt{head}_{S_0} / 2^\texttt{precision}\rfloor$ (where subscripts denote the coder state in which we evaluate a variable).
Since $S_0$ satisfies invariant~(i) by assumption, we have $\texttt{head}_{S_0} < 2^{2\times\texttt{precision}}$, and so $\texttt{head}_{S'} < 2^\texttt{precision}$, which (temporarily) violates invariant~(ii).
This means that the condition for \texttt{conditional\_refill} is satisfied, and it is easy to see that Line~\ref*{line:ans_transfer_pop} inverts Lines~\ref*{line:ans_transfer_push_begin}-\ref*{line:ans_transfer_push_end}.
Thus, Scenario~(\ref*{item:appendix_correctness_scenario_pushpop}) restores the \texttt{AnsCoder} into its original state~$S_0$.

\paragraph{Scenario~(\ref*{item:appendix_correctness_scenario_poppush}):}
Calling \texttt{pop} before \texttt{push} leads to the folowing state transition:
\begin{align}
    S_0 \xrightarrow{\;\texttt{pop\_from\_head}\;}
    S_1 \xrightarrow{\;\texttt{conditional\_refill}\;}
    S_2 \xrightarrow{\;\texttt{conditional\_flush}\;}
    S_3 \xrightarrow{\;\texttt{push\_onto\_head}\;}
    S_4.
\end{align}
We show that state $S_4=S_0$.
Since \texttt{push\_onto\_head} is the inverse of \texttt{pop\_from\_head}, it suffices to show that $S_3 = S_1$.
Note that we can assume the invariants only about~$S_0$.
The intermediate state~$S_1$ may violate invariant~(ii), but it is easy to see that it satisfies invariant~(i):
Lines~\ref*{line:ans_regular_pop_begin}-\ref*{line:ans_regular_pop_end} set
\begin{align}\label{eq:appendix_correctness_scenario_b1}
    \texttt{head}_{S_1} = \left\lfloor \frac{\texttt{head}_{S_0}}{n} \right\rfloor \times m_i(x_i) + z_i'
\end{align}
with $z_i' = z_i - \sum_{x_i'<x_i} m_i(x_i')$ where $z_i = (\texttt{head}_{S_0} \,\text{mod}\, n)$ is the value that gets initialized on Line~\ref*{line:ans_regular_pop_begin}.
Since Lines~\ref*{line:ans_find_symbol_begin}-\ref*{line:ans_find_symbol_end} find the \texttt{symbol}~$x_i$ that satisfies $z_i \in\ZZ_i(x_i)$, we have $z_i < \sum_{x_i' \leq x_i} m_i(x_i')$ (see Eq.~\ref{eq:def-subrange} of the main text) and thus $z_i' < m_i(x_i)$.
Further, we have $\texttt{head}_{S_0} < n^2$ by invariant~(i).
Thus, we find the upper bound
\begin{align}
    \texttt{head}_{S_1} &\leq \left\lfloor \frac{n^2 - 1}{n} \right\rfloor \times m_i(x_i) + (m_i(x_i) -1) \nonumber\\
    &= (n-1) m_i(x_i) + m_i(x_i) - 1 \nonumber\\
    &= n \, m_i(x_i) - 1. \label{eq:appendix_correctness_scenario_b2}
\end{align}
Since the $m_i(x_i')$ are nonnegative for all $x_i'$ and sum to~$n$, we have $m_i(x_i) \leq n$, and so Eq.~\ref{eq:appendix_correctness_scenario_b2} implies $\texttt{head}_{S_1} < n^2$, and thus $S_1$ satisfies invariant~(i).
We now distinguish two cases depending on whether $S_1$ satisfies invariant~(ii).
If $S_1$ satisfies invariant~(ii), then \texttt{conditional\_refill} is a no-op and $S_2=S_1$.
In the next step, \texttt{conditional\_flush} checks whether $\lfloor \texttt{head}_{S_1} / n\rfloor \geq m_i(x_i)$ on Line~\ref*{line:ans_check_invariant_push}, which cannot be the case due to Eq.~\ref{eq:appendix_correctness_scenario_b2}, and thus \texttt{conditional\_flush} is also a no-op and we have $S_3 = S_2 = S_1$.
If, on the other hand, $S_1$ does not satisfy invariant~(ii), then \texttt{bulk} was not empty at the entry of the method \texttt{pop}, and thus the assumption that the original state~$S_0$ satisfies invariant~(ii) implies $\texttt{head}_{S_0} \geq n$.
Line~\ref*{line:ans_transfer_pop} then sets \texttt{head} to the new value $\texttt{head}_{S_2} = \texttt{head}_{S_1} \times n + b$ with some $b\geq 0$.
In the next step, \texttt{conditional\_flush} checks whether $\lfloor \texttt{head}_{S_2} / n\rfloor \geq m_i(x_i)$, which is now indeed the case since $\lfloor \texttt{head}_{S_2} /n \rfloor \geq \lfloor \texttt{head}_{S_1} \times n / n \rfloor = \texttt{head}_{S_1}$, which is at least $m_i(x_i)$ according to Eq.~\ref{eq:appendix_correctness_scenario_b1} since $\texttt{head}_{S_0} \geq n$ in this case.
The flushing operation on Lines~\ref*{line:ans_transfer_push_begin}-\ref*{line:ans_transfer_push_end} is thus executed and it inverts the refilling operation from Line~\ref*{line:ans_transfer_pop}, and we again have $S_3=S_1$, as claimed.
Thus, Scenario~(\ref*{item:appendix_correctness_scenario_poppush}) also restores the \texttt{AnsCoder} into its original state~$S_0$.

\subsection{Proof That Invariants Are Uphold}
\label{sec:appendix_correctness_invariants}

The above prove of correctness relied on the assumption that invariants (i) and~(ii) from Section~\ref{sec:streaming-ans} of the main text always hold at the entry of methods \texttt{push} and \texttt{pop}.
We now show that this assumption is justified by proving that the constructor initializes an \texttt{AnsCoder} into a state that satisfies both invariants, and that all methods uphold the invariants (i.e., both invariants hold on method exit provided that they hold on method entry).

\paragraph{Constructor.}
It is easy to see that both invariants are satisfied when the constructor finishes:
the loop condition on Line~\ref*{line:ans_constructor_establish_invariant1} of Listing~\ref{listing:ans} checks if invariant~(ii) is violated, i.e., if \texttt{bulk} is empty \emph{and} $\texttt{head} < n$ (the latter is equivalent to $\texttt{head} \,{\texttt{>>}\, \texttt{precision} = 0}$ where \texttt{>>} denotes a right bit-shift, i.e., integer division by $n=2^\texttt{precision}$).
The loop thus only terminates once invariant~(ii) is satisfied (it is guaranteed to terminate since the loop body on Line~\ref*{line:ans_constructor_establish_invariant2} \texttt{pop}s a word from \texttt{bulk} and the loop terminates at latest once \texttt{bulk} is empty).
Invariant~(i) is satisfied throughout the constructor since \texttt{head} is initialized as zero (which is smaller than~$n^2$) and the only statement that mutates \texttt{head} in the constructor is Line~\ref*{line:ans_constructor_establish_invariant2}.
This statement is only executed if $\texttt{head}$ has at most \texttt{precision} valid bits, and it increases the number of valid bits by at most \texttt{precision} (due to the left bit-shift~$\texttt{<<}$), so it can never lead to more than $2\times\texttt{precision}$ valid bits, which would be necessary to violate invariant~(i).

\paragraph{Encoding.}
We now analyze the method \texttt{push} in Listing~\ref{listing:ans} of the main text.
Assuming that both invariants hold at method entry, we denote by $h$ and~$h'$ the value of \texttt{head} at method entry and exit, respectively, and we distinguish two cases depending on whether or not the \texttt{if}-condition on Line~\ref*{line:ans_check_invariant_push} is met.

\begin{itemize}
\item
\emph{If the \texttt{if}-condition on Line~\ref*{line:ans_check_invariant_push} is met} then we have
\begin{align}\label{eq:appendix_invariant_push1a}
    h \,\texttt{>>}\, \texttt{precision} \geq m_i(x_i)
    \quad\text{and}\quad
    h' = \left(\left\lfloor \frac{h\,\texttt{>>}\, \texttt{precision}}{m_i(x_i)} \right\rfloor \texttt{<<}\, \texttt{precision}\right) + z_i
\end{align}
with $z_i \in \{0,\ldots,2^\texttt{precision}-1\}$ (which is why the addition in Eq.~\ref{eq:appendix_invariant_push1a} can be implemented as a slightly faster bitwise~``\texttt{or}'' in the code).
Using $n=2^\texttt{precision}$, we can rewrite the right and left bit shifts in Eq.~\ref{eq:appendix_invariant_push1a} as integer division by~$n$ (rounding down) and multiplication with~$n$, respectively.
Thus, Eq.~\ref{eq:appendix_invariant_push1a} simplifies to
\begin{align}\label{eq:appendix_invariant_push1b}
    \left\lfloor \frac{h}{n} \right\rfloor \geq m_i(x_i)
    \qquad\text{and}\qquad
    h' = \left\lfloor \frac{\lfloor h / n \rfloor}{m_i(x_i)} \right\rfloor \times n  + z_i
\end{align}
Combining both parts of Eq.~\ref{eq:appendix_invariant_push1b}, we find
\begin{align}\label{eq:appendix_invariant_push1c}
    h' \geq \left\lfloor \frac{m_i(x_i)}{m_i(x_i)} \right\rfloor \times n + z_i
    \geq n = 2^\texttt{precision}
\end{align}
and thus invariant~(ii) is satisfied at method exit.
To show that invariant~(i) is also satisfied at method exit, we start from the assumption that invariant~(i) holds at method entry, i.e., $h < n^2$.
Therefore, we have $h/n<n$ and thus $\lfloor h/n\rfloor \leq n-1$ (since $n$ is an integer).
Inserting this into the second part of Eq.~\ref{eq:appendix_invariant_push1b} leads to
\begin{align}\label{eq:appendix_invariant_push1d}
    h' \leq \left\lfloor \frac{n-1}{m_i(x_i)} \right\rfloor \times n + z_i
    \stackrel{(*)}{\leq} (n-1)\times n + z_i
    = n^2 - (n-z_i)
    < n^2
\end{align}
where, in the equality marked with ``$(*)$'', we used that $m_i(x_i) > 0$ for the encoded symbol~$x_i$ by assumption (and thus $m_i(x_i) \geq 1$ since $m_i(x_i)$ is an integer).
In the last equality, we used $z_i<n$.
Thus, invariant~(i) is also satisfied at method exit.

\item
\emph{If the \texttt{if}-condition on Line~\ref*{line:ans_check_invariant_push} is not met} we have
\begin{align}\label{eq:appendix_invariant_push2a}
    \left\lfloor \frac{h}{n}\right\rfloor < m_i(x_i)
    \qquad\text{and}\qquad
    h' = \left\lfloor \frac{h}{m_i(x_i)} \right\rfloor \times n + z_i.
\end{align}
We can simplify the first part of Eq.~\ref{eq:appendix_invariant_push2a} to $h/n < m_i(x_i)$ since $m_i(x_i)$ is an integer, and we thus find $h/m_i(x_i) < n$ (using the fact that $m_i(x_i)\neq 0$ by assumption for the encoded symbol~$x_i$) and thus $\lfloor h/m_i(x_i)\rfloor \leq n-1$.
Inserting into the second part of Eq.~\ref{eq:appendix_invariant_push2a} leads to
\begin{align}\label{eq:appendix_invariant_push2c}
    h' \leq (n-1) \times n + z_i = n^2 - (n-z_i) < n^2
\end{align}
where we agin used $z_i<n$ in the final step.
Thus invariant~(i) is satisfied at method exit.
Regarding invariant~(ii), we note that $m_i(x_i)\leq n$ since $\sum_{x_i'\in\XX_i} m_i(x_i')=n$, where all terms in the sum are nonnegative.
Thus, the smallest we can make~$h'$ in Eq.~\ref{eq:appendix_invariant_push2a} is if we set $m_i(x_i)=n$ and $z_i=0$, which results in the lower bound
\begin{align}\label{eq:appendix_invariant_push2b}
    h' \geq \left\lfloor \frac{h}{n} \right\rfloor \times n.
\end{align}
Assuming that invariant~(ii) holds at method entry, we know that either \texttt{bulk} is empty, in which case it remains empty at method exit (since we're considering the case where the \texttt{if}-condition on Line~\ref*{line:ans_check_invariant_push} is not met, in which case we don't mutate \texttt{bulk});
or $h\geq n$ which, combined with Eq.~\ref{eq:appendix_invariant_push2b} implies that $h'>n$.
Thus, either way, invariant~(ii) remains satisfied at method exit.
\end{itemize}

\paragraph{Decoding.}
We finally analyze the method \texttt{pop} in Listing~\ref{listing:ans} of the main text.
We show that calling \texttt{pop(m)} upholds invariants~(i) and~(ii) (regardless of whether or not it follows a call of \texttt{push(symbol,~m)} with the same~\texttt{m}).
We denote the value of \texttt{head} at entry of the method \texttt{push} by capital letter~$H$ so that there is no confusion with the discussion of \texttt{push}.
Once the program flow reaches Line~\ref*{line:ans_check_invariant_pop}, \texttt{head} takes the (intermediate) value
\begin{align}\label{eq:appendix_invariant_pop1}
    \tilde H = \left\lfloor \frac{H}{n} \right\rfloor \times m_i(x_i) + z_i'
\end{align}
where $z_i'\in\{0,\ldots,m_i(x_i)-1\}$ and $x_i$ is the decoded symbol, which implies that $\ZZ_i(x_i)$ is not empty and therefore $m_i(x_i) = |\ZZ_i(x_i)| \geq 1$.
It is easy to see that $\tilde H<n^2$, i.e., $\tilde H$~satisfies invariant~(i) since the largest we can make $\tilde H$ in Eq.~\ref{eq:appendix_invariant_pop1} is by setting $H=n^2-1$ (the largest value allowed by invariant~(i)), $m_i(x_i) = n$, and $z_i' = n-1$.
This leads to the upper bound
\begin{align}
    \tilde H \leq \left\lfloor \frac{n^2-1}{n} \right\rfloor \times n + (n-1)
    = (n-1) \times n + (n-1) = n^2-1 < n^2.
\end{align}
The \texttt{if}-condition on Line~\ref*{line:ans_check_invariant_pop} checks if invariant~(ii) is violated.
We again distinguish two cases:
\begin{itemize}
\item
\emph{If invariant~(ii) holds on Line~\ref*{line:ans_check_invariant_pop}} then the method exits, and thus both invariants hold on method exit.

\item
\emph{If invariant~(ii) does not hold on Line~\ref*{line:ans_check_invariant_pop}} then the \texttt{if}-branch is taken and Line~\ref*{line:ans_transfer_pop} mutates \texttt{head} into the final state
\begin{align}\label{eq:appendix_invariant_pop3}
    H' = \tilde H \times n + b
\end{align}
where $b\in\{0,\ldots,n-1\}$ is the word of data that was transferred from \texttt{bulk}.
We can see directly that $H'$~satisfies invariant~(i) since, as $\tilde H$ violates invariant~(ii) we have $\tilde H \leq n-1$ and therefore $H' \leq (n-1) \times n + b = n^2 + (b-n) < n^2$ since $b<n$.
Regarding invariant~(ii), we first note that invariant~(ii) can only be violated on Line~\ref*{line:ans_check_invariant_pop} if \texttt{bulk} at method entry is nonempty.
Thus, the assumption that invariant~(ii) is satisfied at method entry implies that $H \geq n$.
This allows us to see that $\tilde H\neq 0$ since both $\lfloor H/n\rfloor$ and $m_i(x_i)$ are nonzero in Eq.~\ref{eq:appendix_invariant_pop1}.
Inserting into Eq.~\ref{eq:appendix_invariant_pop3} leads to $H' \geq n$, i.e., invariant~(ii) is also satisfied at method exit.
\end{itemize}

In summary, we proved that the \texttt{AnsCoder} from Listing~\ref{listing:ans} of the main text upholds the two invariants stated in Section~\ref{sec:streaming-ans} of the main text.
Using this property, we proved in Section~\ref{sec:appendix_correctness_conditional} that the \texttt{AnsCoder} is a correct encoder/decoder, i.e., that \texttt{pop} is both the left- and the right-inverse of \texttt{push}.

\end{document}